\pdfoutput=1

\documentclass[11pt]{article}

\usepackage{ACL2023}
\usepackage[utf8]{inputenc}
\usepackage{times}
\usepackage{latexsym}

\usepackage{booktabs}       
\usepackage{amsfonts}       
\usepackage{nicefrac}       
\usepackage{xcolor}         
\usepackage{todonotes}
\usepackage{amssymb}
\usepackage{mathtools}
\usepackage{algorithm} 
\usepackage{algpseudocode} 
\usepackage[autostyle]{csquotes}  
\usepackage{hyperref}       
\usepackage{url}  
\usepackage{microtype}
\usepackage{graphicx}
\usepackage[T1]{fontenc}

\usepackage{microtype}

\title{Cross-lingual AMR Aligner: Paying Attention to Cross-Attention}

\author{Abelardo Carlos Mart\'inez Lorenzo$^{1,2*}$ \qquad Pere-Llu\'is Huguet Cabot$^{1,2}$\thanks{$^*$ Equal contributions.} \\
\bf{ Roberto Navigli$^2$}\\
         $^1$ Babelscape, Italy \\
         $^2$ Sapienza NLP Group, Sapienza University of Rome \\
         \texttt{\{martinez,huguetcabot\}@babelscape.com} \\
         \texttt{navigli@diag.uniroma1.it}}

\begin{document}
\maketitle
\begin{abstract}

This paper introduces a novel aligner for Abstract Meaning Representation (AMR) graphs that can scale cross-lingually, and is thus capable of aligning units and spans in sentences of different languages. Our approach leverages modern Transformer-based parsers, which inherently encode alignment information in their cross-attention weights, allowing us to extract this information during parsing. This eliminates the need for English-specific rules or the Expectation Maximization (EM) algorithm that have been used in previous approaches. In addition, we propose a guided supervised method using alignment to further enhance the performance of our aligner. 
We achieve state-of-the-art results in the benchmarks for AMR alignment and demonstrate our aligner's ability to obtain them across multiple languages. Our code will be available at \href{https://www.github.com/Babelscape/AMR-alignment}{github.com/Babelscape/AMR-alignment}.

\end{abstract}

\section{Introduction}
\begin{figure*}[t!]
    \centering
        \includegraphics[width=2\columnwidth]{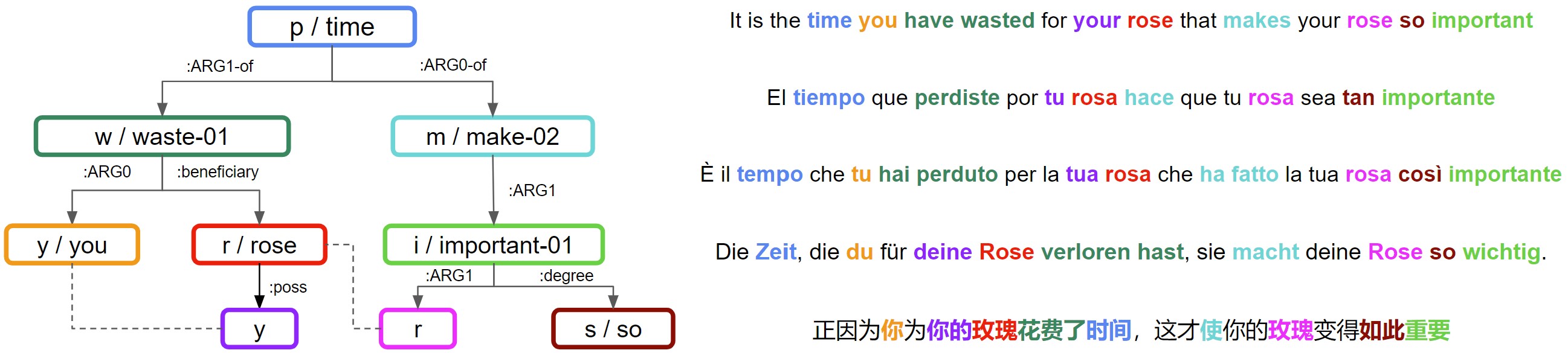}
    \caption{AMR graph (left) and its corresponding sentences in several languages (right). Colors represent alignment.} 
    \label{fig:amr-example}
\end{figure*}

At the core of Natural Language Understanding lies the task of Semantic Parsing, aimed at translating natural language text into machine-interpretable representations. One of the most popular semantic formalisms is the Abstract Meaning Representation \cite[AMR]{banarescu-etal-2013-abstract}, which embeds the semantics of a sentence in a directed acyclic graph, where concepts are represented by nodes, such as \textit{time}, semantic relations between concepts by edges, such as \textit{:beneficiary}, and the co-references by reentrant nodes, such as \textit{r} representing \textit{rose}. In cross-lingual AMR, the English AMR graph represents the sentence in different languages (see Figure \ref{fig:amr-example}). 
To date, AMR has been widely used in  Machine Translation \citep{song-etal-2019-semantic}, Question Answering \citep{lim-etal-2020-know, kapanipathi-etal-2021-leveraging}, Human-Robot Interaction \citep{bonial-etal-2020-dialogue}, Text Summarization \citep{hardy-vlachos-2018-guided,liao-etal-2018-abstract} and Information Extraction \citep{rao-etal-2017-biomedical}, among other areas. 

The alignment between spans in text and semantic units in AMR graphs is an essential requirement for a variety of purposes, including training AMR parsers \citep{zhou-etal-2021-structure}, cross-lingual AMR parsing \citep{blloshmi-etal-2020-xl}, downstream task application \citep{song-etal-2019-semantic}, or the creation of new semantic parsing formalisms \citep{bmr-etal-2022-bmr, martinez-lorenzo-etal-2022-fully}. Despite the emergence of various alignment generation approaches, such as rule-based methods \citep{liu-etal-2018-amr} and statistical strategies utilizing Expectation Maximization (EM) \citep{pourdamghani-etal-2014-aligning, blodgett-schneider-2021-probabilistic}, these methods rely heavily on English-specific rules, making them incompatible with cross-lingual alignment. Furthermore, even though several attempts extend the alignment to non-English sentences and graphs~\cite{damonte-cohen-2018-cross, uhrig-etal-2021-translate}, these efforts are inherently monolingual and therefore lack the connection to the richer AMR graph bank available in English, which can be exploited as a source of interlingual representations.

On the other hand, current state-of-the-art AMR parsers are auto-regressive neural models \citep{bevilacqua-etal-2021-one, amrbart-2022-acl} that do not generate alignment when parsing the sentence to produce the graph. Therefore, to obtain both, one needs to i) predict the graph and then ii) generate the alignment using an aligner system that is based on language-specific rules.

This paper presents the first AMR aligner that can scale cross-lingually by leveraging the implicit information acquired in Transformer-based parsers \citep{amrbart-2022-acl}. We propose an approach for extracting alignment information from cross-attention, and a guided supervised method to enhance the performance of our aligner. We eliminate the need for language-specific rules and enable simultaneous generation of the AMR graph and alignment. Our approach is efficient and robust, and is suitable for cross-lingual alignment of AMR graphs.

Our main contributions are: (i) we explore how Transformer-based AMR parsers preserve implicit alignment knowledge and how we can extract it; (ii) we propose a supervised method using cross-attention to enhance the performance of our aligner, (iii) we achieve state-of-the-art results along different alignment standards and demonstrate the effectiveness of our aligner across languages.

\section{Related Work}

\paragraph{AMR alignment}
Since the appearance of AMR as a Semantic Parsing formalism, several aligner systems have surfaced that provide a link between the sentence and graph units. JAMR \citep{flanigan-etal-2014-discriminative} is a widely used aligner system that employs an ordered list of 14 criteria, including exact and fuzzy matching, to align spans to subgraphs. However, this approach has limitations as it is unable to resolve ambiguities or learn novel alignment patterns. TAMR \citep{liu-etal-2018-amr} extends JAMR by incorporating an oracle parser that selects the alignment corresponding to the highest-scored candidate AMR graph. ISI \citep{pourdamghani-etal-2014-aligning} aligner utilizes an EM algorithm to establish alignment between words and graphs' semantic units. First, the graph is linearized, and then the EM algorithm is employed with a symmetrized scoring function to establish alignments. This method leads to more diversity in terms of alignment patterns, but fails to align easy-to-recognize patterns that could be aligned using rules. LEAMR \citep{blodgett-schneider-2021-probabilistic} is another aligner system that combines rules and EM. This approach aligns all the subgraph structures to any span in the sentence. However, it is based on language-specific rules, making it unsuitable for cross-lingual settings. Moreover, despite several attempts to extend the alignment to non-English languages \citet{anchieta-pardo-2020-semantically, oral-eryigit-2022-amr}, these efforts are still monolingual since they rely on language-specific strategies. Consequently, in this paper we present an approach that fills this gap.

\paragraph{Cross-attention}
Most state-of-the-art systems for AMR parsing are based on Encoder-Decoder Transformers, specifically on BART~\cite{lewis-etal-2020-bart}. These models consist of two stacks of Transformer layers, which utilize self- and cross-attention as their backbone. The popularity of Transformer models has led to increased interest in understanding how attention encodes information in text and relates to human intuition~\cite{Vashishth2019AttentionIA} and definitions of explainability ~\cite{bastings-filippova-2020-elephant,bibal2022attention}. Research has been conducted on how attention operates, relates to preconceived ideas, aggregates information, and explains model behavior for tasks such as natural language inference\cite{stacey2022aaai}, Translation~\cite{yin-etal-2021-context, zhang-feng-2021-modeling-concentrated, chen-etal-2021-mask}, Summarization~\cite{xu-etal-2020-self, manakul-gales-2021-long} or Sentiment Analysis~\cite{wu-etal-2020-structured}. Furthermore, there have been attempts to guide attention to improve interpretability or performance in downstream tasks~\cite{deshpande-narasimhan-2020-guiding, NEURIPS2020_460191c7}. However, to the best of our knowledge, there has been no prior study on attention for AMR parsing. This paper fills this gap by investigating the role of attention in AMR parsing.

\section{Method}
\label{sec:method}

Originally described by \citet{NIPS2017_3f5ee243} as \enquote{multi-head attention over the output of the Encoder}, and referred to as cross-attention in \citet{lewis-etal-2020-bart}, it enables the Decoder to attend to the output of the Encoder stack, conditioning the hidden states of the autoregressive component on the input text. Self-attention and cross-attention modules are defined as:
$$\operatorname{Attention}(Q, K, V) = \operatorname{att}(Q,K)V $$
$$ \operatorname{att}(Q,K)= softmax(\frac{QK^T}{\sqrt{d_k}})$$
$$\operatorname{CrossAtt}(Q, K, V) =$$ 
$$ \operatorname{Concat}(head_1, ..., head_H)W^O$$ 
$$ head_h = \operatorname{Attention}(Q W^Q_h, K W^K_h, V W^V_h)$$

\noindent where $K, V = E^\ell \in \mathbb{R}^{n_e \times d_kH}$ and $Q = D^\ell \in \mathbb{R}^{n_d \times d_kH}$ are the Encoder and Decoder hidden states at layer $\ell$, $n_e$ and $n_d$ are the input and output sequence lengths, $H$ is the number of heads,  $W^Q_h, W^K_h$ and $W^V_h \in \mathbb{R}^{d_kH \times d_k}$ are learned weights that project the hidden states to the appropriate dimensions, $d_k$, for each head and $W^O \in \mathbb{R}^{d_kH \times d_kH}$ is a final learned linear projection.
Therefore in each head $h$ and layer $\ell$ we define the attention weights as 
 $att_h^\ell = \operatorname{att}(D^\ell W^Q_{h},E^\ell W^K_h) \in \mathbb{R}^{n_d \times n_e}$. 

\subsection{Unguided Cross-Attention}
We argue that there is an intuitive connection between cross-attention and alignments. Under the assumption the Decoder will attend to the parts in the input that are more relevant to predicting the next token, we infer that, when decoding the tokens for a certain node in the graph, attention should focus on related tokens in the input, and therefore the words that align to that node.
We will use the cross-attention matrices ($att_h^\ell$) 
to compute an alignment between the input and the output.

\subsection{Guided Cross-Attention}
\label{sec:supervised}

We also aim to explore whether cross-attention can be guided by the alignment between the words of the sentence and the nodes of the graph. To this end, we construct a sparse matrix $align \in \mathbb{R}^{n_d \times n_e}$ from the automatically-generated alignments:

$$ align(i,j) = \left\{ \begin{array}{lcc}
             1 &   if  & x_i \sim y_j \\
             \\ 0 &  if & x_i \nsim y_j \\
             \end{array}
   \right. $$
   
\noindent where $\sim$ indicates alignment between subword token $x_i$ and graph token $y_j$.

However, even though there are sparse versions of attention~\cite{10.5555/3045390.3045561}, these did not produce successful alignments in our experiments. Hence we choose to alleviate the constraint of imposing sparsity by employing the scalar mixing approach introduced in ELMo~\cite{peters-etal-2018-deep}. We learn a weighted mix of each head and obtain a single attention matrix:
\begin{equation}
\label{eq:weight_mix}
    att^\ell = \gamma \sum^{H-1}_{h=0} s_h^\ell att_h^\ell \in \mathbb{R}^{n_d \times n_e}
\end{equation}

where $\textbf{s} = softmax(\textbf{a})$ with scalar learnable parameters $\gamma, a_0, \dots, a_H$.

The model has the flexibility to learn how to distribute weights such that certain heads give sparser attention similar to alignment, while others can encode additional information that is not dependent on alignment. In our experiments, we use the implementation of \citet[SPRING]{bevilacqua-etal-2021-one} to train our parser but add an extra cross-entropy loss signal:
\begin{multline*}
     \mathcal{L} = -\sum_{j=1}^{n_d}\log{p_{BART}\left(y_{j} \mid y_{<j}, x\right)} \\ -\sum_{\mathclap{\substack{j=1  \\ ~~~~~~~~~~\sum_i align(i,j) > 0}}}^{n_d} \sum_{i=1}^{n_e} \log\left({\frac{e^{att^\ell(i,j)}}{ \sum\limits_{k=1}^{n_d}  e^{att^\ell(i,k)}} \frac{align(i,j)}{\sum\limits^{n_e}_{k=1} align(k,j)}}\right)
\end{multline*}

\subsection{Saliency Methods}\label{sec:saliency}

A theoretical alternative to our reasoning about cross-attention is the use of input saliency methods.
These methods assign higher importance to the input tokens that correspond to a particular node in the graph or were more important in their prediction during decoding. To obtain these importance weights, we employ Captum~\cite{kokhlikyan2020captum}, an open-source library for model interpretability and understanding, which provides a variety of saliency methods, including gradient-based methods such as Integrated Gradients (IG), Saliency~\cite{Simonyan2014DeepIC}, and Input X Gradient (IxG), backpropagation-based methods such as Deeplift~\cite{10.5555/3305890.3306006} and Guided Backpropagation (GB)~\cite{Springenberg2015StrivingFS}, and finally occlusion-based methods~\cite{Zeiler2014VisualizingAU}.

We obtain a weight matrix $sal \in \mathbb{R}^{n_d \times n_e}$ with the same size as the cross-attention matrix and use it to extract alignments in the same fashion as the unguided cross-attention method. This approach allows us to explore the input tokens that have a greater impact on the decoding process and can aid in understanding the reasoning behind the alignments made by the model.

\subsection{Alignment Extraction}
\label{sec:align_ext}

Our algorithm\footnote{The pseudo-algorithm is described in the Appendix \ref{sec:alignment-extraction-algorithm}.} to extract and align the input-output spans is divided into six steps:

\begin{enumerate}
    \item \textbf{Alignment score matrix:} we create a matrix $M \in \mathbb{R}^{n_d \times n_e}$, where $n_e$ is the number of tokens in the sentence and $n_d$ is the number of tokens in the linearized graph, using the cross-attention 
    weights ($att_h^\ell$ or $att^\ell$) as described in Section \ref{sec:method}.
    \item \textbf{Span segmentation:} For each sentence word, we sum the scores of tokens that belong to the same word  column-wise in $M$. Then, for LEAMR  alignments (see Section \ref{sec:standards}), the sentence tokens are grouped into spans using their span segmentation (see Appendix \ref{sec:rules}).
    \item \textbf{Graph segmentation:} We sum the score of tokens that belong to the same graph's semantic unit row-wise in $M$. 
    \item \textbf{Sentence graph tokens map:} We iterate over all the graph's semantic units and map them to the sentence span with highest score in $M$.
    \item \textbf{Special graph structures:} We revise the mapping by identifying subgraphs that represent literal or matching spans -- e.g., named entities, dates, specific predicates, etc. -- and align them accordingly.
    \item \textbf{Alignment formatting:} We extract the final alignments to the appropriate format using the resulting mapping relating graph's semantic units to sentence spans. 
\end{enumerate}

\begin{figure*}[t!]
    \centering
    \def\svgwidth{0.99\columnwidth}
    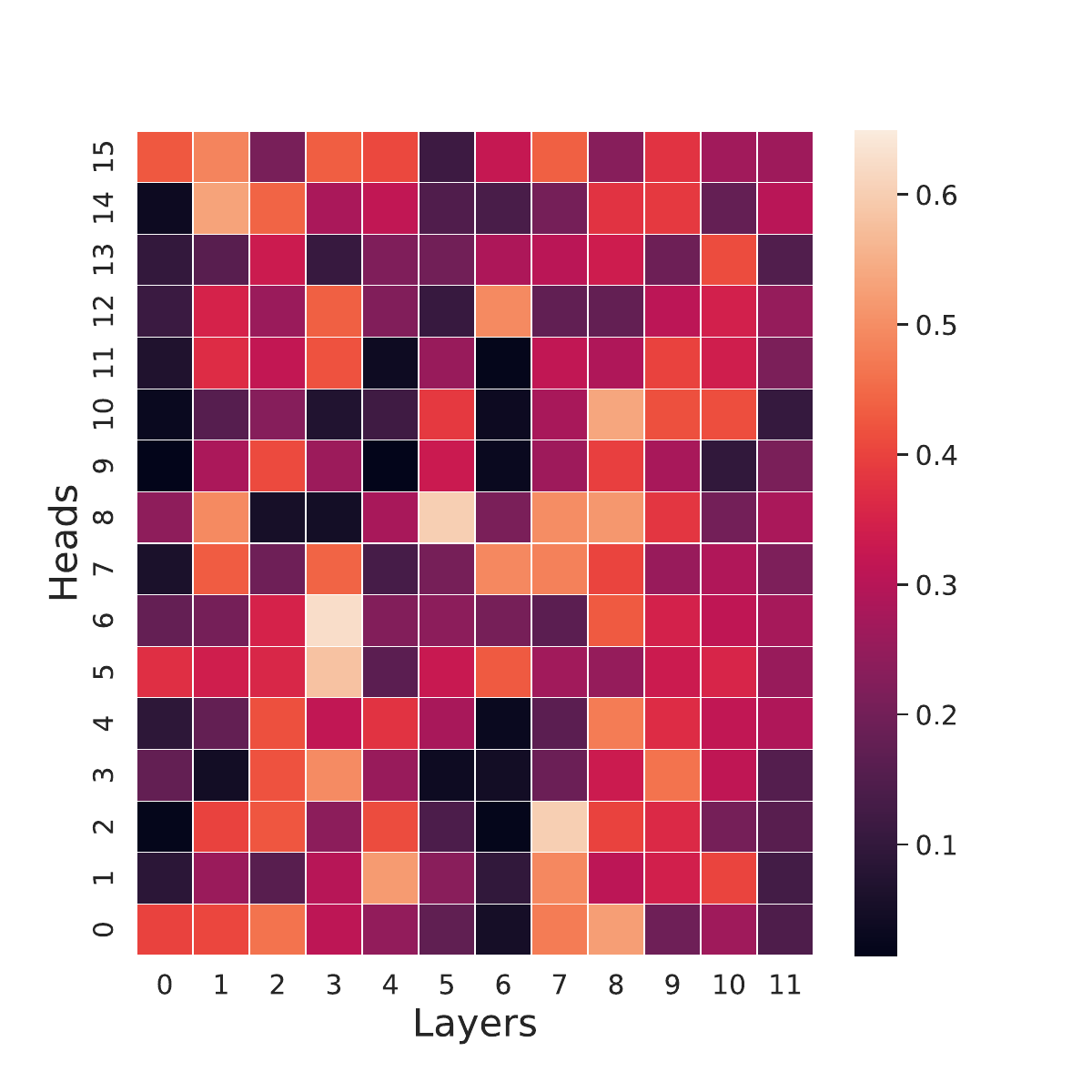
    \def\svgwidth{0.99\columnwidth}
    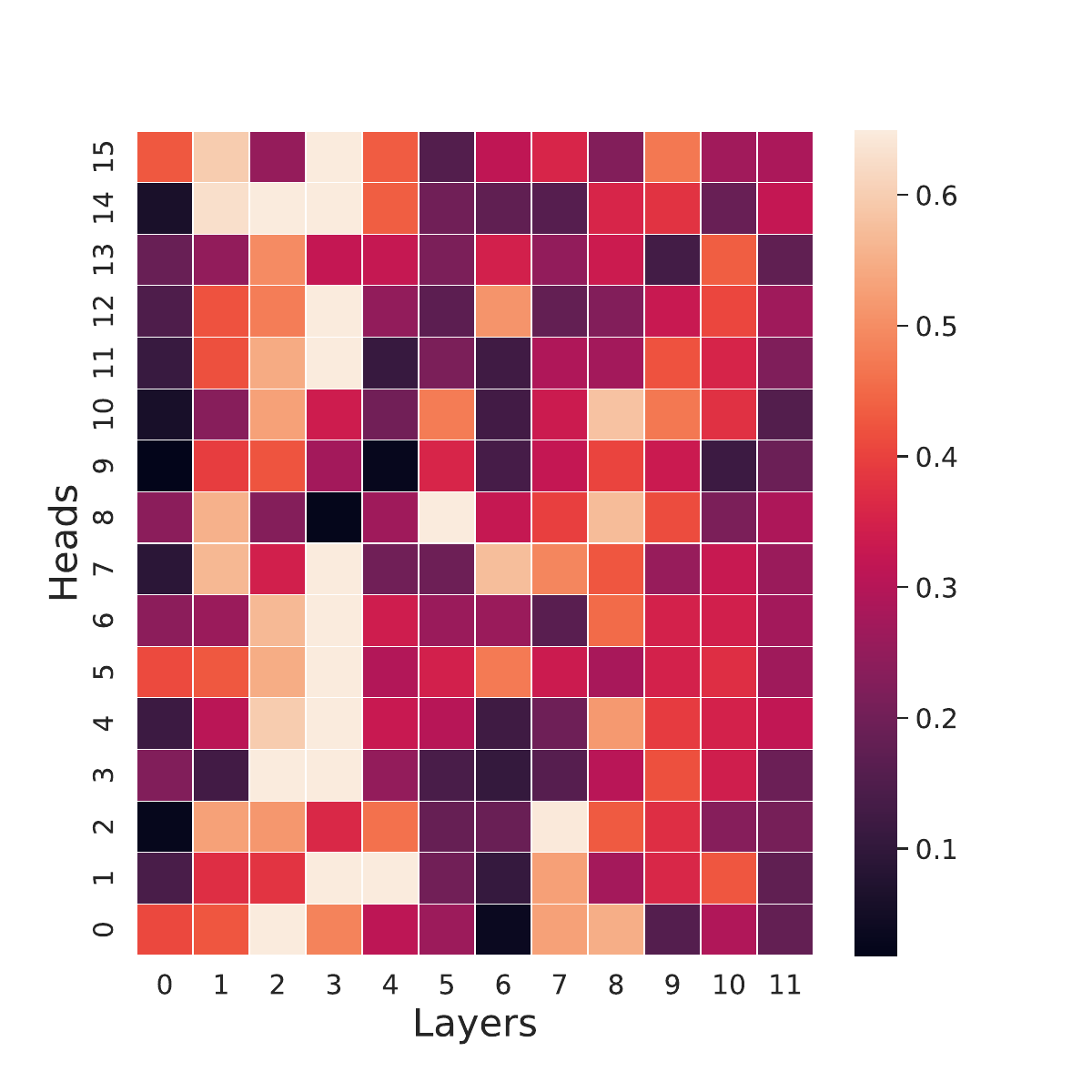
    \def\svgwidth{\columnwidth}
    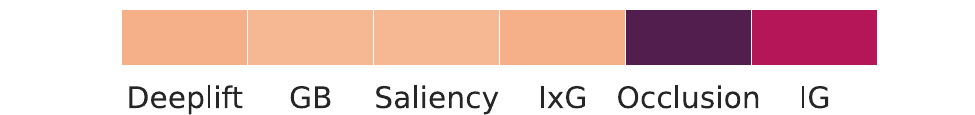
    \caption{Heatmap of Pearson's r correlation to LEAMR validation set for unguided (left) and guided on half the heads in layer 3 (right) cross-attention weights, as well as saliency methods (bottom).}
    \label{fig:dev_corr}
\end{figure*}

\section{Experimental Setup}

\subsection{Graph inventory} AMR 3.0 (LDC2020T02) consists of 59,255 sentence-graph pairs that are manually annotated. However, it lacks alignment information between the nodes in the graphs and the concepts in the sentences. We use the train split for the guided approach and use the respective validation and test splits from the alignment systems. Additionally, to evaluate cross-lingual performance, we use the gold German, Italian, and Spanish sentences of ``AMR 2.0 -- Four Translation'' (LDC2020T07) which are human parallel translations of the test set in AMR 2.0\footnote{The sentences of AMR 2.0 are a subset of AMR 3.0.}, paired with their English graphs from the AMR 3.0 test set. Despite this, as the graph inventory does not contain alignment information, it becomes necessary to access other repositories in order to obtain the alignment.

\subsection{Alignment Standards} \label{sec:standards}
We propose an approach that is agnostic to different alignment standards and we evaluate it on two standards that are commonly used: ISI and LEAMR. 

\paragraph{ISI} The ISI standard, as described in~\cite{pourdamghani-etal-2014-aligning}, aligns single spans in the sentence to graphs' semantic units (nodes or relations), and aligns relations and reentrant nodes when they appear explicitly in the sentence. The alignments are split into two sets of 200 annotations each, which we use as validation and test sets, updated to the AMR 3.0 formalism. For the cross-lingual alignment setup, we project English ISI graph-sentence alignments to the sentences in other languages, using the machine translation aligner~\cite{dou-neubig-2021-word}. This involves connecting the nodes in the graph to the spans in non-English sentences using the projected machine translation alignments between the English spans and the corresponding non-English sentence spans. By leveraging this, we are able to generate a silver alignment for cross-lingual AMR, which enables us to validate the model's performance in a cross-lingual setup and determine its scalability across-languages.

\paragraph{LEAMR} The LEAMR standard differentiates among four different types of alignment: i) Subgraph Alignments, where all the subgraphs that explicitly appear in the sentence are aligned to a list of consecutive spans, ii) Duplicate Subgraph, where all the subgraphs that represent omit repeated concepts in the sentence are aligned, iii) Relation Alignments, where all the relations that were not part of a previous subgraph structure are aligned, and iv) Reentrancy Alignments, where all the reentrant nodes are aligned. In contrast to ISI, all the semantic units in the graph are aligned to some list of consecutive spans in the text. We use 150 alignments as the validation set and 200 as the test set, which includes sentence-graph pairs from \textit{The Little Prince} Corpus (TLP) complemented with randomly sampled pairs from AMR 3.0.

\subsection{Model}
\label{sec:model}
We use SPRING~\cite{bevilacqua-etal-2021-one} as our parsing model based on the BART-large architecture~\cite{lewis-etal-2020-bart} for English and SPRING based on mBART for non-English languages  mBART~\cite{liu-etal-2020-multilingual-denoising} for the multilingual setting. We extract all $att_h^\ell$ matrices from a model trained on AMR 3.0 as in \citet{blloshmi-etal-2021-spring} in order to perform our unguided cross-attention analysis. For the guided approach we re-train using the same hyperparameters as the original implementation, but with an extra loss signal as described in Section \ref{sec:supervised} based on either LEAMR or ISI. When using LEAMR alignments, we restructure the training split in order to exclude any pair from their test and validation sets.

\section{Experiments}

\subsection{Correlation} \label{sec:experiments}

\begin{figure*}[t!]
    \centering
    \includegraphics[width=2\columnwidth]{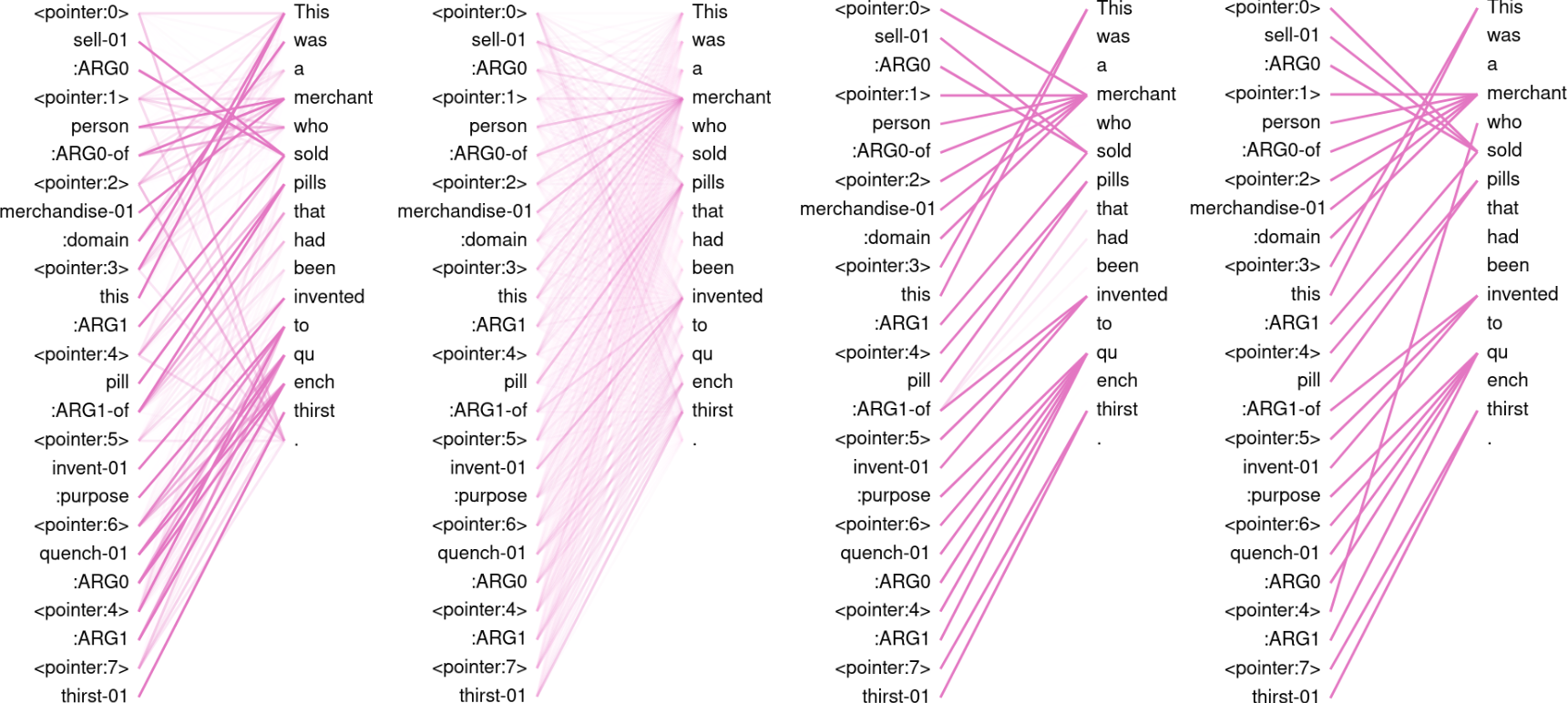}
    \caption{Unguided (left), saliency (center-left) and guided (center-right) alignment weights and LEAMR (right) gold alignment for lpp\_1943.1209. To explore all cross-
attention weights interactively, please go \href{https://amr-align-cross-attention.netlify.app/}{here}.}
    \label{fig:example}
\end{figure*}

In this study, we investigate the correlation between cross-attention and alignment by computing the Pearson's r correlation coefficient between the $att_h^\ell$ matrix and the LEAMR alignment matrix $align$. To do so, we first flatten the matrices and remove any special tokens that are not relevant for alignment. As shown in Figure \ref{fig:dev_corr}, there is a clear positive correlation between the two.

While we do not have a clear explanation for why certain heads have a higher correlation than others, it is evident that there is a connection between cross-attention and alignment. For example, head 6 in layer 3 (i.e., $att_6^3$) has a correlation coefficient of 0.635, approximately the same as the sum of the entire layer.

With regard to the saliency methods described in Section \ref{sec:saliency}, the two most highly correlated methods were Saliency and GB, with a correlation coefficient of 0.575. Despite this result, we observe that saliency methods tend to focus more on essential parts of the sentence, such as the subject or predicate. These parts are usually aligned to more nodes and relations, which explains the high correlation, but they lack nuance compared to cross-attention.

Our best results were obtained by supervising layer 3 during training with the approach outlined in Section \ref{sec:supervised}, using Cross-Entropy Loss on half of the heads (i.e., 3, 4, 5, 6, 7, 11, 12, and 15) that were selected based on their correlation on the validation set. This did not affect the performance of parsing. When we looked at $att^3$ using the learned weighted mix from Equation \ref{eq:weight_mix} with LEAMR alignments, the correlation reached 0.866, which is significantly higher than any other method. Figure \ref{fig:dev_corr} shows the impact of supervising half the heads on layer 3 and how it influences heads in other layers.

To gain a better understanding of these results, we present an example from the TLP corpus in Figure \ref{fig:example} to illustrate the different methods, including cross-attention and saliency methods. The left image shows the cross-attention values for $att_6^3$. Despite not having seen any alignment information, the model is able to correctly match non-trivial concepts such as "merchant" and "person". The center-left image illustrates how saliency methods focus on essential parts of the sentence, but lack nuance compared to cross-attention. The center-right image shows that supervising learning on layer 3 results in more condensed attention, which is associated with the improvement in correlation. However, it is important to note that the model can reliably attend to incorrect positions, such as aligning "pointer" to "merchant" instead of "sold".

\subsection{Results} \label{sec:results}

\paragraph{LEAMR} Table \ref{tab:leamr-experiment} shows the performances of our two approaches on the LEAMR gold alignments compared to previous systems. We use the same evaluation setup as \citet{blodgett-schneider-2021-probabilistic}, where the partial match assigns a partial credit from Jaccard indices between nodes and tokens. 
In both guided and unguided methods, we extract the score matrix for Algorithm \ref{alg:pseudo} from the sum of the cross-attention in the first four layers.  We use a Wilcoxon signed-rank test~\cite{10.2307/3001968} on the alignment matches per graph to check for significant differences. Both our approaches are significantly different compared to LEAMR (p=0.031 and p=0.007 respectively). However, we find no statistical difference between  our unguided and guided approaches (p=0.481).

Our guided attention approach performs best, improving upon LEAMR on Subgraph (+0.5) and Relation (+2.6). 
For Reentrancy, performance is relatively low, and we will explore the reasons for this in Section \ref{sec:errors}. Perhaps most interesting is the performance of the unguided system using raw cross-attention weights from SPRING. The system remains competitive against the guided model without having access to any alignment information. It outperforms LEAMR which, despite being unsupervised with respect to alignments, relies on a set of inductive biases and rules based on alignments. While we also draw on specific rules related to the graph structure in post-processing, we will need to investigate their impact in an ablation study.

\begin{table*}[t]
\centering
\resizebox{\textwidth}{!}{%
\begin{tabular}{ll|ccc|ccc|c|c}
\toprule
\textbf{}                                                        & 
\multicolumn{1}{l|}{\textbf{}}                 & \multicolumn{3}{c|}{\textbf{Exact Alignment}}       & \multicolumn{3}{c|}{\textbf{Partial Alignment}} & \multicolumn{1}{l|}{\textbf{Spans}} & \multicolumn{1}{l}{\textbf{Coverage}} \\
\textbf{}                                         & \multicolumn{1}{l|}{\textbf{}}                 & \textbf{P}                    & \textbf{R}                                   & \multicolumn{1}{c|}{\textbf{F1}}     & \textbf{P}                    & \textbf{R}                                     & \multicolumn{1}{c|}{\textbf{F1}}     & \multicolumn{1}{c|}{\textbf{F1}}     & \textbf{}                             \\ \midrule \midrule
\multicolumn{1}{l|}{\textbf{Subgraph}}            & \multicolumn{1}{l|}{ISI}                       & 71.56                         & 68.24                                        & \multicolumn{1}{c|}{69.86}          & 78.03                         & 74.54                                          & \multicolumn{1}{c|}{76.24}          & \multicolumn{1}{c|}{86.59}          & \phantom{0}78.70                                  \\
\multicolumn{1}{l|}{\textbf{Alignment}}    & \multicolumn{1}{l|}{JAMR}                      & 87.21                         & 83.06                                        & \multicolumn{1}{c|}{85.09}          & 90.29                         & 85.99                                          & \multicolumn{1}{c|}{88.09}          & \multicolumn{1}{c|}{92.38}          & \phantom{0}91.10                                  \\
\multicolumn{1}{l|}{(1707)}                    & \multicolumn{1}{l|}{TAMR}                      & 85.68                         & 83.38                                        & \multicolumn{1}{c|}{84.51}          & 88.62                         & 86.24                                          & \multicolumn{1}{c|}{87.41}          & \multicolumn{1}{c|}{94.64}          & \phantom{0}94.90                                  \\
\multicolumn{1}{l|}{}                             & \multicolumn{1}{l|}{LEAMR}               & 93.91                         & 94.02                                        & \multicolumn{1}{c|}{93.97}          & 95.69                         & 95.81                                          & \multicolumn{1}{c|}{95.75}          & \multicolumn{1}{c|}{96.05}          & 100.00                                  \\
\multicolumn{1}{l|}{\textbf{}}                    & \multicolumn{1}{l|}{LEAMR $\dagger$}                & 93.74                         & 93.91                                        & \multicolumn{1}{c|}{93.82}          & 95.51                         & 95.68                                          & \multicolumn{1}{c|}{95.60}          & \multicolumn{1}{c|}{95.54}          & 100.00                                   \\
\multicolumn{1}{l|}{}                             & \multicolumn{1}{l|}{Ours - Unguided}       & 94.11                & 94.49                               & \multicolumn{1}{c|}{94.30} & 96.03                         & 96.42                                 & \multicolumn{1}{c|}{96.26}          & \multicolumn{1}{c|}{95.94}          & 100.00                                   \\
\multicolumn{1}{l|}{}                             & \multicolumn{1}{l|}{Ours - Guided - ISI}   & 89.87                         & 91.97                                        & \multicolumn{1}{c|}{90.91}          & 92.11                         & 94.27                                          & \multicolumn{1}{c|}{93.18}          & \multicolumn{1}{c|}{93.69}          & 100.00                                   \\
\multicolumn{1}{l|}{}                             & \multicolumn{1}{l|}{Ours - Guided - LEAMR} & \textbf{94.39}                         & \textbf{94.67}                                        & \multicolumn{1}{c|}{\textbf{94.53}}          & \textbf{96.62}                & \textbf{96.90}                                          & \multicolumn{1}{c|}{\textbf{96.76}} & \multicolumn{1}{c|}{96.40}          & 100.00                                   \\ \midrule \midrule
\multicolumn{1}{l|}{\textbf{Relation}}           & \multicolumn{1}{l|}{ISI}                       & 59.28                         & \phantom{0}8.51                                         & \multicolumn{1}{c|}{14.89}          & 66.32                         & \phantom{0}9.52                                           & \multicolumn{1}{c|}{16.65}          & \multicolumn{1}{c|}{83.09}          & \phantom{00}9.80                                   \\
\multicolumn{1}{l|}{\textbf{Alignment}}             & \multicolumn{1}{l|}{LEAMR}               & 85.67                         & 87.37                                        & \multicolumn{1}{c|}{85.52}          & 88.74                         & 88.44                                          & \multicolumn{1}{c|}{88.59}          & \multicolumn{1}{c|}{95.41}          & 100.00                                   \\
\multicolumn{1}{l|}{(1263)}                             & \multicolumn{1}{l|}{LEAMR $\dagger$}                & 84.63                         & 84.85                                        & \multicolumn{1}{c|}{84.74}          & 87.77                         & 87.99                                          & \multicolumn{1}{c|}{87.88}          & \multicolumn{1}{c|}{91.98}          & 100.00                                   \\
\multicolumn{1}{l|}{}                             & \multicolumn{1}{l|}{Ours - Unguided}       & 87.14                         & 87.59                               & \multicolumn{1}{c|}{87.36} & 89.87                         & 90.33                                 & \multicolumn{1}{c|}{90.10} & \multicolumn{1}{c|}{91.03}          & 100.00                          \\
\multicolumn{1}{l|}{}                             & \multicolumn{1}{l|}{Ours - Guided - ISI}   & 83.82                         & 83.39                                        & \multicolumn{1}{c|}{83.61}          & 86.45                        & 86.00                                          & \multicolumn{1}{c|}{86.22}          & \multicolumn{1}{c|}{87.30}          & 100.00                                   \\
\multicolumn{1}{l|}{}                             & \multicolumn{1}{l|}{Ours - Guided - LEAMR} & \textbf{88.03}                & \textbf{88.18}                                       & \multicolumn{1}{c|}{\textbf{88.11}}          & \textbf{91.08}                & \textbf{91.24}                                          & \multicolumn{1}{c|}{\textbf{91.16}}          & \multicolumn{1}{c|}{91.87}          & 100.00                                   \\ \midrule \midrule
\multicolumn{1}{l|}{\textbf{Reentrancy}}           & \multicolumn{1}{l|}{LEAMR}               & 55.75                         & 54.61                                       & \multicolumn{1}{c|}{55.17}          &         ---                      &                  ---                              &  ---             &  ---              & 100.00                                   \\
\multicolumn{1}{l|}{\textbf{Alignment}}             & \multicolumn{1}{l|}{LEAMR $\dagger$}                & 54.61                         & 54.05                               & \multicolumn{1}{c|}{54.33} &     ---                          &            ---                                    &  ---            & ---              & 100.00                                   \\
\multicolumn{1}{l|}{(293)}                             & \multicolumn{1}{l|}{Ours - Unguided}       & 44.75                         & 44.59                                        & \multicolumn{1}{c|}{44.67}          &     ---                          &             ---                                   &  ---             & ---              & 100.00                                   \\
\multicolumn{1}{l|}{}                             & \multicolumn{1}{l|}{Ours - Guided - ISI}   & 42.09                         & 39.35                                        & \multicolumn{1}{c|}{40.77}          &        ---                       &             ---                                   &  ---               & ---               & 100.00                                   \\
\multicolumn{1}{l|}{}                             & \multicolumn{1}{l|}{Ours - Guided - LEAMR} & \textbf{56.90}                & \textbf{57.09}                                        & \multicolumn{1}{c|}{\textbf{57.00}}          & ---                     & ---                                      & \multicolumn{1}{c|}{---}      & \multicolumn{1}{c|}{---}      & 100.00                                   \\ \midrule \midrule
\multicolumn{1}{l|}{\textbf{Duplicate}} & \multicolumn{1}{l|}{LEAMR}               & 66.67                         & 58.82                                        & \multicolumn{1}{c|}{62.50}           & 70.00                            & 61.76                                          & \multicolumn{1}{c|}{65.62}          & ---         & 100.00                                   \\
\multicolumn{1}{l|}{\textbf{Subgraph}}               & \multicolumn{1}{l|}{LEAMR $\dagger$}                & 68.75                         & 64.71                                        & \multicolumn{1}{c|}{66.67}          & 68.75                         & 64.71                                          & \multicolumn{1}{c|}{66.67}          & ---         & 100.00                                   \\
\multicolumn{1}{l|}{\textbf{Alignment}}                             & \multicolumn{1}{l|}{Ours - Unguided}       & \textbf{77.78}                & \textbf{82.35}                                        & \multicolumn{1}{c|}{\textbf{80.00}}             & \textbf{77.78}                & \textbf{82.35}                                          & \multicolumn{1}{c|}{\textbf{80.00}}             & ---        & 100.00                                   \\
\multicolumn{1}{l|}{(17)}                             & \multicolumn{1}{l|}{Ours - Guided - ISI} & 63.16                            & 70.59                                        & \multicolumn{1}{c|}{66.67}          & 65.79                         & 73.53                                          & \multicolumn{1}{c|}{69.44}          & ---               & 100.00                                   \\
\multicolumn{1}{l|}{}                             & \multicolumn{1}{l|}{Ours - Guided - LEAMR}                     & 70.00                            & 82.35                               & 75.68                     & 72.50                            & 85.29                                 & 78.38                    & ---                           & 100.00                                   \\ \bottomrule
\end{tabular}%
}
\caption{LEAMR alignment results. Column blocks: models; Exact and Partial alignment scores; Span and Coverage measures. Row blocks: alignment types, number of instances in brackets. $\dagger$ indicates our re-implementation. Guided versions using ISI/LEAMR silver alignments.
Bold is best.}
\label{tab:leamr-experiment}
\end{table*}

\begin{table}[h!]
\centering
\resizebox{\columnwidth}{!}{%
\begin{tabular}{l|l|ccc}
\toprule
\textbf{}          & \textbf{AMR parser}  & \textbf{P}                        & \textbf{R}                        & \textbf{F1}                       \\ \midrule \midrule
\textbf{ALL}       & ISI                  & 59.3                              & 08.5                              & 14.9                              \\
                   & LEAMR $\dagger$             & 84.6                              & 84.9                              & 84.7                              \\
                   & Ours - Unguided  & 87.1                              & 87.6                              & 87.4                              \\
                   & Ous - Guided - LEAMR & \textbf{88.0}                     & \textbf{88.2}                     & \textbf{88.1}                     \\ \midrule \midrule
\textbf{Single}    & ISI                  & \textbf{82.9}                     & 52.1                              & 64.0                              \\
\textbf{Relations} & LEAMR   $\dagger$             & 64.8                              & 55.7                              & 59.9                              \\
\textbf{(121)}     & Ours - Unguided  & 79.5                              & \textbf{79.5}                     & \textbf{79.5}                     \\
                   & Ous - Guided - LEAMR & 77.5                              & 64.8                              & 70.5                              \\ \midrule \midrule
\textbf{Argument}  & ISI                  & 39.6                              & 03.5                              & 06.4                              \\
\textbf{Structure} & LEAMR   $\dagger$             & 86.6                              & 88.2                              & 87.4                              \\
\textbf{(1042)}    & Ours - Unguided  & \multicolumn{1}{l}{87.9}          & \multicolumn{1}{l}{88.4}          & \multicolumn{1}{l}{88.2}          \\
                   & Ous - Guided - LEAMR & \multicolumn{1}{l}{\textbf{89.0}} & \multicolumn{1}{l}{\textbf{90.8}} & \multicolumn{1}{l}{\textbf{89.9}} \\ \bottomrule
\end{tabular}%
}
\caption{LEAMR results breakdown for Relation Alignment. Column blocks: relation type; models; scores. Bold is best. $\dagger$ indicates our re-implementation.}
\label{tab:results-relation}
\end{table}

\begin{table*}[ht]
\centering
\resizebox{\textwidth}{!}{%
\begin{tabular}{l|rrr|rrr|rrr|rrr|rrr}
\toprule
& \multicolumn{1}{c}{\textbf{}}  & \multicolumn{1}{c}{\textbf{EN}} & \multicolumn{1}{c|}{\textbf{}}    & \multicolumn{1}{c}{\textbf{}}  & \multicolumn{1}{c}{\textbf{DE}} & \multicolumn{1}{c|}{\textbf{}}       & \multicolumn{1}{c}{\textbf{}}  & \multicolumn{1}{c}{\textbf{ES}} & \multicolumn{1}{c|}{\textbf{}}      & \multicolumn{1}{c}{\textbf{}}  & \multicolumn{1}{c}{\textbf{IT}} & \multicolumn{1}{c|}{\textbf{}}      & \multicolumn{1}{c}{}  & \multicolumn{1}{c}{\textbf{AVG}} & \multicolumn{1}{c}{}   \\ 
\multicolumn{1}{l|}{}                      & \multicolumn{1}{c}{\textbf{P}} & \multicolumn{1}{c}{\textbf{R}}  & \multicolumn{1}{c|}{\textbf{F1}} & \multicolumn{1}{c}{\textbf{P}} & \multicolumn{1}{c}{\textbf{R}}  & \multicolumn{1}{c|}{\textbf{F1}}    & \multicolumn{1}{c}{\textbf{P}} & \multicolumn{1}{c}{\textbf{R}}  & \multicolumn{1}{c|}{\textbf{F1}}   & \multicolumn{1}{c}{\textbf{P}} & \multicolumn{1}{c}{\textbf{R}}  & \multicolumn{1}{c|}{\textbf{F1}}   & \multicolumn{1}{c}{\textbf{P}} & \multicolumn{1}{c}{\textbf{R}}   & \multicolumn{1}{c}{\textbf{F1}} \\ \midrule \midrule
\multicolumn{1}{l|}{\textbf{JAMR}}         & 92.7                           & 80.1                            & \multicolumn{1}{r|}{85.9}        & \textbf{75.4}                  & ~6.6                            & \multicolumn{1}{r|}{12.1}           & \textbf{84.4}                  & 16.1                            & \multicolumn{1}{r|}{27.1}          & 64.8                           & 13.2                            & \multicolumn{1}{r|}{21.9}          & \textbf{79.3}         & 29.0                    & 36.8                   \\
\multicolumn{1}{l|}{\textbf{TAMR}}         & 92.1                           & 84.5                            & \multicolumn{1}{r|}{88.1}        & 73.7                           & ~6.4                            & \multicolumn{1}{r|}{11.8}           & 84.0                           & 16.4                            & \multicolumn{1}{r|}{27.5}          & 64.3                           & 13.2                            & \multicolumn{1}{r|}{21.9}          & 78.5                  & 30.1                    & 37.3                   \\
\multicolumn{1}{l|}{\textbf{LEAMR}}        & 85.9                           & 92.3                            & \multicolumn{1}{r|}{89.0}        & ~8.4                           & ~9.3                            & \multicolumn{1}{r|}{~8.8}           & ~8.1                           & ~9.0                            & \multicolumn{1}{r|}{~8.5}          & ~9.0                           & ~9.5                            & \multicolumn{1}{r|}{~9.3}          & 27.9                  & 30.0                    & 28.9                   \\
\multicolumn{1}{l|}{\textbf{Unguided}} & 95.4                           & 93.2                            & \multicolumn{1}{r|}{94.3}        & 64.0                           & \textbf{74.4}                   & \multicolumn{1}{r|}{\textbf{68.85}} & 67.9                           & \textbf{77.1}                   & \multicolumn{1}{r|}{\textbf{72.2}} & \textbf{67.4}                  & \textbf{75.5}                   & \multicolumn{1}{r|}{\textbf{71.2}} & 73.7                  & \textbf{80.1}           & \textbf{76.6}          \\
\textbf{Guided}                            & \textbf{96.3}                  & \textbf{94.2}                   & \textbf{95.2}                    & \multicolumn{1}{c}{---}          & \multicolumn{1}{c}{---}           & \multicolumn{1}{c|}{---}               & \multicolumn{1}{c}{---}          & \multicolumn{1}{c}{---}           & \multicolumn{1}{c|}{---}              & \multicolumn{1}{c}{---}          & \multicolumn{1}{c}{---}           & \multicolumn{1}{c|}{---}              & \multicolumn{1}{c}{---}  & \multicolumn{1}{c}{---}    & \multicolumn{1}{c}{---}   \\ \bottomrule

\end{tabular}%
}
\caption{ISI results. Column blocks: models, language.}
\label{tab:isi-experiment}
\end{table*}

\begin{table*}[t!]
\centering
\resizebox{\textwidth}{!}{%
\begin{tabular}{l|ccc||ccc||ccccccccccc}
\toprule
               & \multicolumn{3}{c||}{\textbf{GOLD}}                                  & \multicolumn{3}{c||}{\textbf{Without Rules}}                               & \multicolumn{11}{c}{\textbf{Layers}}                                                                                                                                 \\ 
               & \multicolumn{1}{l}{} & \multicolumn{1}{l}{} & \multicolumn{1}{l||}{} & \multicolumn{1}{l}{} & \multicolumn{1}{l}{} & \multicolumn{1}{l||}{} & \multicolumn{5}{c}{\textbf{Unguided}}                                                        & \multicolumn{6}{c}{\textbf{Guided}}                                      \\
               & LEAMR $\dagger$                & Ung.                 & Guided                & LEAMR $\dagger$                & Ung.                 & Guided                & Sal.           & {[}0:4{]}     & {[}4:8{]} & {[}8:12{]} & \multicolumn{1}{c||}{{[}0:12{]}} & {[}0:4{]}     & {[}4:8{]} & {[}8:12{]} & {[}0:12{]} & {[}3{]} & {[}3{]}* \\ \midrule \midrule
\textbf{Sub.}  & 96.5                 & 96.7                 & \textbf{97.0}         & 87.6                 & 88.6                 & \textbf{93.4}         & 62.2           & \textbf{94.3} & 69.8      & 63.3       & \multicolumn{1}{c||}{87.7}       & \textbf{94.5} & 74.4      & 66.3       & 93.2       & 93.7    & 93.7     \\
\textbf{Rel.}  & 87.1                 & 89.2                 & \textbf{90.3}         & 26.6                 & 60.1                 & \textbf{83.4}         & 50.0           & \textbf{87.7} & 72.7      & 61.6       & \multicolumn{1}{c||}{84.5}       & \textbf{88.1} & 73.8      & 62.5       & 87.9       & 86.2    & 85.9     \\
\textbf{Reen.} & 56.8                 & 46.7                 & \textbf{59.0}         & 15.2                 & 38.6                 & \textbf{57.0}         & 34.5           & \textbf{44.7} & 41.1      & 36.1       & \multicolumn{1}{c||}{41.9}       & \textbf{57.0} & 39.2      & 33.0       & 51.0       & 52.7    & 53.4     \\
\textbf{Dupl.} & 62.9                 & \textbf{80.0}        & 75.7                  & 40.0                 & 71.8                 & \textbf{73.7}         & \phantom{0}9.5 & \textbf{80.0} & 11.1      & 27.3       & \multicolumn{1}{c||}{64.3}       & \textbf{75.9} & 30.0      & 27.3       & 66.7       & 70.3    & 70.3     \\ \hline
\end{tabular}
}
\caption{F1 results on Exact Alignment on ablation studies. Column blocks: alignment types; using gold spans; removing rules from the models; by layers. Guided approach using LEAMR silver alignments. $\dagger$ indicates our re-implementation. [x:y] indicates sum from layer x to y. * indicates weighted head sum. Bold is best.}
\label{tab:ablations}
\end{table*}

Relations that are argument structures (i.e., :$ARG$ and :$ARG$-\textit{of}) usually depend on the predictions for their parent or child nodes; hence their improvement would be expected to be tied to the Subgraph Alignment. The results in Table \ref{tab:results-relation} reassure us that this intuition is correct. Notice how for Single Relations (such as :$domain$ or :$purpose$ in Figure \ref{fig:example}) the performance by LEAMR was much lower, even worse than that of ISI:  \citet{blodgett-schneider-2021-probabilistic} argued that this was due to the model being overeager to align to frequent prepositions such as \textit{to} and \textit{of}. On the other hand, our unguided method achieves 15 points over ISI and 20 over LEAMR, which hints at the implicit knowledge on alignment that cross-attention encodes. 
Our guided approach experiences a considerable drop for Single Relations since it was trained on data generated by LEAMR, replicating its faulty behavior albeit being slightly more robust.

\paragraph{ISI} When we test our systems against the ISI alignments, both our models achieve state-of-the-art results, surpassing those of previous systems, including LEAMR. This highlights the flexibility of cross-attention as a standard-agnostic aligner (we provide additional information in Appendix \ref{sec:appendix_isi}).
Table \ref{tab:isi-experiment} shows the performance of our systems and compares ones with the ISI alignment as a reference. We omit relations and Named Entities to focus solely on non-rule-based alignments and have a fair comparison between systems. Here, our aligner does not rely on any span-segmentation, hence nodes and spans are aligned solely based on which words and nodes share the highest cross-attention values. Still, both our alignments outperform those of the comparison systems in English. 

Moreover, only our approach achieves competitive results in Spanish, German and Italian -- obtaining 40 points more on average above the second best model -- while the other approaches are hampered by the use of English-specific rules. However, we found two reasons why  non-English systems perform worse than in English: 1) linguistic divergences (as explained in \citet{wein-schneider-2021-classifying}), and ii) the machine translation alignment error. 

\section{Ablation Study}
\label{sec:ablation}

\textbf{Gold spans}$\quad$ LEAMR relies on a span segmentation phase, with a set of multiword expressions and Stanza-based named entity recognition. We use the same system in order to have matching sentence spans. However, these sometimes differ from the gold spans, leading to errors. Table \ref{tab:ablations} (left) shows performance using an oracle that provides gold spans, demonstrating how our approach still outperforms LEAMR across all categories.

\textbf{Rules}$\quad$ All modern alignment systems depend on rules to some degree. For instance, we use the subgraph structure for Named Entities, certain relations are matched to their parent or child nodes, etc. (see Appendix \ref{sec:rules} for more details). But what is the impact of such rules? As expected, both LEAMR and our unguided method see a considerable performance drop when we remove them. For Relation, LEAMR drops by almost 60 points, since it relies heavily on the predictions of parent and child nodes to provide candidates to the EM model. Our unguided approach also suffers from such dependency, losing 25 points. However, our guided model is resilient to rule removal, dropping by barely one point on Subgraph and 5 points on Relation.

\textbf{Layers}$\quad$ Figure \ref{fig:dev_corr} shows how alignment acts differently across heads and layers. We explore this information flow in the Decoder by extracting the alignments from the sum of layers at different depths. The right of Table \ref{tab:ablations} shows this for both our unguided and guided models, as well as the Saliency method. [3] indicates the sum of heads in the supervised layer, while [3]* is the learned weighted mix. From our results early layers seem to align more explicitly, with performance dropping with depth. This corroborates the idea that Transformer models encode basic semantic information early~\cite{tenney-etal-2019-bert}. While layers 7 and 8 did show high correlation values, the cross-attention becomes more disperse with depth, probably due to each token encoding more contextual information.

\section{Error Analysis}
\label{sec:errors}
We identify two main classes of error that undermine the extraction of alignments.

\textbf{Consecutive spans} $\quad$ 
Because each subgraph in LEAMR is aligned to a list of successive spans, the standard cannot deal correctly with transitive phrasal verbs. For example, for the verb "take off" the direct object might appear in-between ("He took his jacket off in Málaga"). Because these are not consecutive spans, we align just to "take" or "off".

\textbf{Rules} $\quad$ We have a few rules for recognizing subgraph structures, such as Named Entities, and align them to the same spans. However, Named Entity structures contain a placeholder node indicating the entity type;  when the placeholder node appears explicitly in the sentence, the node should not be part of the Named Entity subgraph. For example, when aligning  \textit{`Málaga', the city}, the placeholder node should be aligned to \textit{city} while our model aligned it to \textit{Málaga}.

\begin{figure}[t!]
    \centering
    \includegraphics[width=0.8\columnwidth]{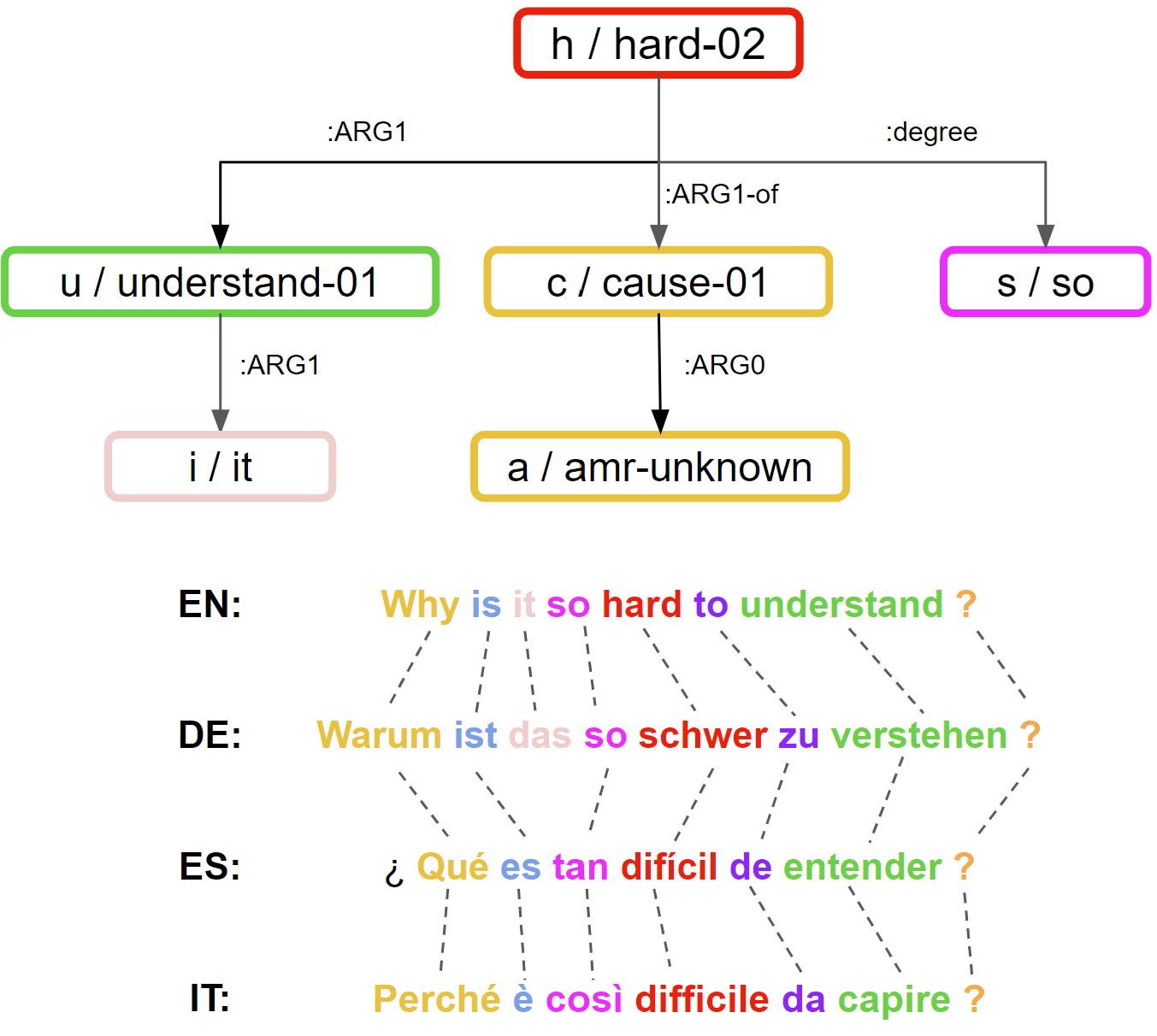}
    \caption{AMR graph and its sentence from \textit{"AMR 2.0 – Four Translation”}. Color represents the alignment.}
    \label{fig:amr-analisy-translations}
\end{figure}

\section{Cross-lingual Analysis: A Case Study}

To investigate the potential causes of misalignment between English and non-English languages, we conduct a case study that qualitatively examines the differences in alignment generated by different systems and languages. Figure \ref{fig:amr-analisy-translations} illustrates the sentence, \textit{"why is it so hard to understand?"} with its human translations in German, Spanish, and Italian, and its AMR. In the Italian translation, the subject of the verb is omitted, while in the Spanish translation the focus of the question is modified from asking the reason \textit{why something is difficult to understand} to asking directly \textit{what is difficult to understand}, making "qué" the subject. As a consequence, in both cases making it impossible to align "it" with any word in either the Italian or Spanish sentence by Machine Translation Alignment. Table \ref{tab:cross-alignment-analisys} presents the alignments generated for the sentence in each language and with each model in ISI format. Although our model was able to align the node \textit{"it"} by aligning it with the conjugated verb in the Italian sentence and with the word \textit{"qué"} in the Spanish sentence, which serves as the subject, this resulted in an error in our evaluation since the alignment of "it" was not projected in either Italian or Spanish. In addition, we also observed the performance of jamr and tamr, which are rule-based systems, and found that they were only able to align the word "so" in the German translation, as it shares the same lemma in English. In contrast, LEAMR was able to detect more alignments due to its requirement to align all nodes to a corresponding word in the target language. However, the alignments generated by LEAMR appeared to be almost entirely random.

\begin{table}[t!]
\centering
\resizebox{\columnwidth}{!}{%
\begin{tabular}{@{}|l|l|l|l|l|l|l|l|l|l|@{}}
\toprule
En       & Why           & is    & it    & so        & hard      & to    & understand  & ?     \\ \hline
ref       & 1.2   1.2.1   & ---   & 1.1.1 & 1.3       & 1         & ---   & 1.1         & ---   \\ 
ours       & 1.2   1.2.1   & ---   & 1.1.1 & 1.3       & 1         & ---   & 1.1         & ---   \\ 
jamr        & ---           & ---   & 1.1.1 & 1.3       & 1         & ---   & 1.1         & ---   \\ 
tamr     & ---           & ---   & 1.1.1 & 1.3       & 1   1.2   & ---   & 1.1         & ---   \\ 
leamr     & 1.2           & 1.1.1 & 1.1.1 & 1.3       & 1         & ---   & 1.1         & 1.2.1 \\  \hline \hline
De         & Warum         & ist   & das   & so        & schwer    & zu    & verstehen   & ?     \\ \hline
ref        & 1.2    1.2.1  & ---   & 1.1.1 & 1.3       & 1         & ---   & 1.1         & ---   \\ 
ours       & 1.2   1.2.1   &       & 1.1.1 & 1.3       & 1         & ---   & 1.1         & ---   \\ 
jamr        & ---           & ---   & ---   & 1.3       & ---       & ---   & ---         & ---   \\ 
tamr       & ---           & ---   & ---   & 1.3       & ---       & ---   & ---         & ---   \\ 
leamr     & 1             & ---   & 1.2   & 1.2       & 1.3       & 1.1.1 & 1.2.1       & ---   \\  \hline \hline
Es          & Qué          & es   & ---     & tan       & díficil   & de    & entender    & ?     \\ \hline
ref      & 1.2   1.2.1   & ---   & ---   & 1.3       & 1         & ---   & 1.1         & ---   \\ 
ours     & 1.1.1   1.2.1 & ---   & ---   & 1.2   1.3 & 1         & ---   & 1.1         & ---   \\ 
jamr    & ---           & ---   & ---   & ---       & ---       & ---   & ---         & ---   \\ 
tamr      & ---           & ---   & ---   & ---       & ---       & ---   & ---         & ---   \\ 
leamr & 1.1.1   1  & 1.2 & ---    & ---       & 1.1       & ---   & 1.3         & 1.2.1 \\ \hline \hline
It      & Perché    & é        & ---    & cosí      & difficile & da    & capire      & ?     \\ \hline
ref       & 1.2   1.2.1   & ---   & ---   & 1.3       & 1         & ---   & 1.1         & ---   \\ 
ours      & 1.2   1.2.1 & 1.1.1   & ---   &  1.3       & 1         & ---   & 1.1         & ---   \\ 
jamr     & ---           & ---   & ---   & ---       & ---       & ---   & ---         & ---   \\ 
tamr      & ---           & ---   & ---   & ---       & ---       & ---   & ---         & ---   \\ 
leamr    & 1           & 1.1    & ---    & 1.3       & 1.1.1     &       & 1.2.1   1.2 &       \\ \bottomrule
\end{tabular}%
}
\caption{Alignments between sentences and graph from Figure \ref{fig:amr-analisy-translations} across diferent system. "ref" is the reference alignment obtained by Machine Translation Aligment.}
\label{tab:cross-alignment-analisys}
\end{table}

\section{Conclusion}
\label{sec:conclusion}

In this paper, we have presented the first AMR aligner that can scale cross-lingually and demonstrated how cross-attention is closely tied to alignment in AMR Parsing. Our approach outperforms previous aligners in English, being the first to align cross-lingual AMR graphs. We leverage the cross-attention from current AMR parsers, without overhead computation or affecting parsing quality. Moreover, our approach is more resilient to the lack of handcrafted rules, highlighting its capability as a standard- and language-agnostic aligner, paving the way for further NLP tasks. As a future direction, we aim to conduct an analysis of the attention heads that are not correlated with the alignment information in order to identify the type of information they capture, such as predicate identification, semantic relations, and other factors. Additionally, we plan to investigate how the alignment information is captured across different NLP tasks and languages in the cross-attention mechanism of sequence-to-sequence models.
Such analysis can provide insights into the inner workings of the models and improve our understanding of how to enhance their performance in cross-lingual settings.

\section{Limitations}

Despite the promising results achieved by our proposed method, there are certain limitations that need to be noted. Firstly, our approach relies heavily on the use of Transformer models, which can be computationally expensive to train and run. Additionally, the lower performance of our aligner for languages other than English is still a substantial shortcoming, which is discussed in Section \ref{sec:results}.

Furthermore, our method is not adaptable to non-Transformer architectures, as it relies on the specific properties of Transformer-based models to extract alignment information. 

Lastly, our method is based on the assumption that the decoder will attend to those input tokens that are more relevant to predicting the next one. However, this assumption may not always hold true in practice, which could lead to suboptimal alignments.

In conclusion, while our proposed method presents a promising approach for cross-lingual AMR alignment, it is important to consider the aforementioned limitations when applying our method to real-world scenarios. Future research could focus on addressing these limitations and exploring ways to improve the performance of our aligner for languages other than English.

\section{Ethics Statement}

While our approach has shown itself to be effective in aligning units and spans in sentences of different languages, it is important to consider the ethical and social implications of our work.

One potential concern is the use of Transformer-based models, which have been shown to perpetuate societal biases present in the data used for training. Our approach relies on the use of these models, and it is  therefore  crucial to ensure that the data used for training is diverse and unbiased. Furthermore, the use of cross-attention in our approach could introduce new ways to supervise a model in order to produce harmful or unwanted model predictions. Therefore, it is crucial to consider the ethical implications of any guidance or supervision applied to models and to ensure that any training data used to guide the model is unbiased and does not perpetuate harmful stereotypes or discrimination.

Additionally, it is important to consider the potential impact of our work on under-resourced languages. While our approach has shown to be effective in aligning units and spans in sentences of different languages, it is important to note that the performance gap for languages other than English still exists. Further research is needed to ensure that our approach is accessible and beneficial for under-resourced languages.

\section*{Acknowledgments}
\begin{center}
\noindent
\begin{minipage}{0.1\linewidth}
    \raisebox{-0.25\height}{\includegraphics[trim =0mm 5mm 5mm 5mm,clip,scale=0.045]{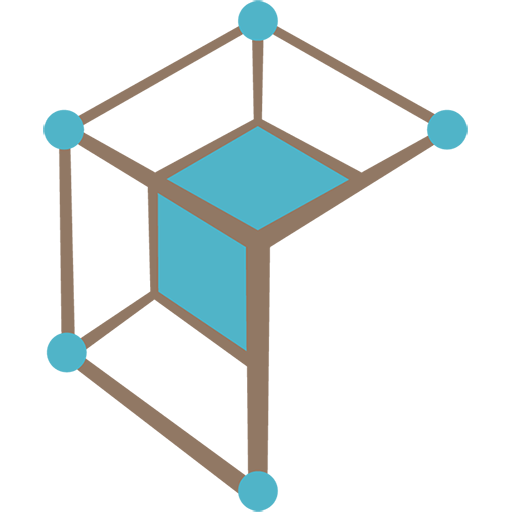}}

\end{minipage}
\hspace{0.005\linewidth}
\begin{minipage}{0.72\linewidth}
The authors gratefully acknowledge the support of the European Union’s Horizon 2020 research project \textit{Knowledge Graphs at Scale} (KnowGraphs) under the Marie  Marie Sk\l{}odowska-Curie grant agreement No \href{https://cordis.europa.eu/project/id/860801}{860801}.

  \vspace{1ex}
\end{minipage}
\hspace{0.005\linewidth}
\begin{minipage}{0.1\linewidth}
  \vspace{0.05cm}
\raisebox{-0.25\height}{\includegraphics[trim =0mm 5mm 5mm 5mm,clip,scale=0.060]{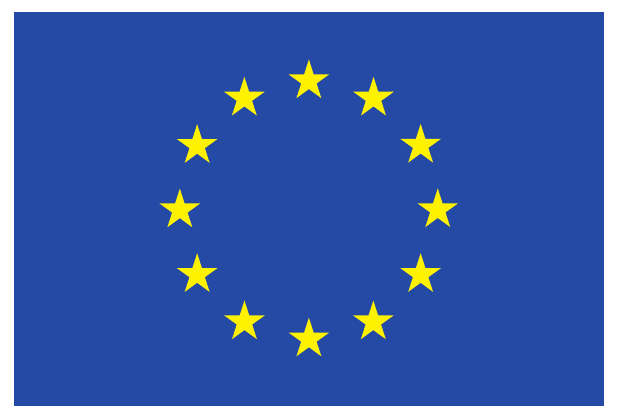}}
  \vspace{0.05cm}
\end{minipage}\\
\end{center}
The last author gratefully acknowledges the support of the PNRR MUR project PE0000013-FAIR.

\bibliography{anthology,custom}

\begin{thebibliography}{49}
\expandafter\ifx\csname natexlab\endcsname\relax\def\natexlab#1{#1}\fi

\bibitem[{Anchi{\^e}ta and Pardo(2020)}]{anchieta-pardo-2020-semantically}
Rafael Anchi{\^e}ta and Thiago Pardo. 2020.
\newblock \href {https://doi.org/10.18653/v1/2020.emnlp-main.123} {Semantically
  inspired {AMR} alignment for the {P}ortuguese language}.
\newblock In \emph{Proceedings of the 2020 Conference on Empirical Methods in
  Natural Language Processing (EMNLP)}, pages 1595--1600, Online. Association
  for Computational Linguistics.

\bibitem[{Bai et~al.(2022)Bai, Chen, and Zhang}]{amrbart-2022-acl}
Xuefeng Bai, Yulong Chen, and Yue Zhang. 2022.
\newblock \href {https://aclanthology.org/2022.acl-long.415} {Graph
  pre-training for {AMR} parsing and generation}.
\newblock In \emph{Proceedings of the 60th Annual Meeting of the Association
  for Computational Linguistics (Volume 1: Long Papers)}, pages 6001--6015,
  Dublin, Ireland. Association for Computational Linguistics.

\bibitem[{Banarescu et~al.(2013)Banarescu, Bonial, Cai, Georgescu, Griffitt,
  Hermjakob, Knight, Koehn, Palmer, and
  Schneider}]{banarescu-etal-2013-abstract}
Laura Banarescu, Claire Bonial, Shu Cai, Madalina Georgescu, Kira Griffitt, Ulf
  Hermjakob, Kevin Knight, Philipp Koehn, Martha Palmer, and Nathan Schneider.
  2013.
\newblock \href {https://aclanthology.org/W13-2322} {{A}bstract {M}eaning
  {R}epresentation for sembanking}.
\newblock In \emph{Proceedings of the 7th Linguistic Annotation Workshop and
  Interoperability with Discourse}, pages 178--186, Sofia, Bulgaria.
  Association for Computational Linguistics.

\bibitem[{Bastings and Filippova(2020)}]{bastings-filippova-2020-elephant}
Jasmijn Bastings and Katja Filippova. 2020.
\newblock \href {https://doi.org/10.18653/v1/2020.blackboxnlp-1.14} {The
  elephant in the interpretability room: Why use attention as explanation when
  we have saliency methods?}
\newblock In \emph{Proceedings of the Third BlackboxNLP Workshop on Analyzing
  and Interpreting Neural Networks for NLP}, pages 149--155, Online.
  Association for Computational Linguistics.

\bibitem[{Bevilacqua et~al.(2021)Bevilacqua, Blloshmi, and
  Navigli}]{bevilacqua-etal-2021-one}
Michele Bevilacqua, Rexhina Blloshmi, and Roberto Navigli. 2021.
\newblock \href
  {https://ojs.aaai.org/index.php/AAAI/article/download/17489/17296} {One
  {SPRING} to rule them both: {S}ymmetric {AMR} semantic parsing and generation
  without a complex pipeline}.
\newblock In \emph{Proceedings of AAAI}.

\bibitem[{Bibal et~al.(2022)Bibal, Cardon, Alfter, Souza~Wilkens, Wang,
  Fran{\c{c}}ois, and Watrin}]{bibal2022attention}
Adrien Bibal, R{\'e}mi Cardon, David Alfter, Rodrigo Souza~Wilkens, Xiaoou
  Wang, Thomas Fran{\c{c}}ois, and Patrick Watrin. 2022.
\newblock \href {https://xiaoouwang.com/xowang/Is_Attention_Explanation.pdf}
  {Is attention explanation? an introduction to the debate}.
\newblock In \emph{Association for Computational Linguistics. Annual Meeting.
  Conference Proceedings}.

\bibitem[{Blloshmi et~al.(2021)Blloshmi, Bevilacqua, Fabiano, Caruso, and
  Navigli}]{blloshmi-etal-2021-spring}
Rexhina Blloshmi, Michele Bevilacqua, Edoardo Fabiano, Valentina Caruso, and
  Roberto Navigli. 2021.
\newblock \href {https://doi.org/10.18653/v1/2021.emnlp-demo.16} {{SPRING}
  {G}oes {O}nline: {E}nd-to-{E}nd {AMR} {P}arsing and {G}eneration}.
\newblock In \emph{Proceedings of the 2021 Conference on Empirical Methods in
  Natural Language Processing: System Demonstrations}, pages 134--142, Online
  and Punta Cana, Dominican Republic. Association for Computational
  Linguistics.

\bibitem[{Blloshmi et~al.(2020)Blloshmi, Tripodi, and
  Navigli}]{blloshmi-etal-2020-xl}
Rexhina Blloshmi, Rocco Tripodi, and Roberto Navigli. 2020.
\newblock \href {https://doi.org/10.18653/v1/2020.emnlp-main.195} {{XL}-{AMR}:
  Enabling cross-lingual {AMR} parsing with transfer learning techniques}.
\newblock In \emph{Proceedings of the 2020 Conference on Empirical Methods in
  Natural Language Processing (EMNLP)}, pages 2487--2500, Online. Association
  for Computational Linguistics.

\bibitem[{Blodgett and Schneider(2021)}]{blodgett-schneider-2021-probabilistic}
Austin Blodgett and Nathan Schneider. 2021.
\newblock \href {https://doi.org/10.18653/v1/2021.acl-long.257} {Probabilistic,
  structure-aware algorithms for improved variety, accuracy, and coverage of
  {AMR} alignments}.
\newblock In \emph{Proceedings of the 59th Annual Meeting of the Association
  for Computational Linguistics and the 11th International Joint Conference on
  Natural Language Processing (Volume 1: Long Papers)}, pages 3310--3321,
  Online. Association for Computational Linguistics.

\bibitem[{Bonial et~al.(2020)Bonial, Donatelli, Abrams, Lukin, Tratz, Marge,
  Artstein, Traum, and Voss}]{bonial-etal-2020-dialogue}
Claire Bonial, Lucia Donatelli, Mitchell Abrams, Stephanie~M. Lukin, Stephen
  Tratz, Matthew Marge, Ron Artstein, David Traum, and Clare Voss. 2020.
\newblock \href {https://aclanthology.org/2020.lrec-1.86} {Dialogue-{AMR}:
  {A}bstract {M}eaning {R}epresentation for dialogue}.
\newblock In \emph{Proceedings of the 12th Language Resources and Evaluation
  Conference}, pages 684--695, Marseille, France. European Language Resources
  Association.

\bibitem[{Chen et~al.(2021)Chen, Sun, and Liu}]{chen-etal-2021-mask}
Chi Chen, Maosong Sun, and Yang Liu. 2021.
\newblock \href {https://doi.org/10.18653/v1/2021.acl-long.369} {Mask-align:
  Self-supervised neural word alignment}.
\newblock In \emph{Proceedings of the 59th Annual Meeting of the Association
  for Computational Linguistics and the 11th International Joint Conference on
  Natural Language Processing (Volume 1: Long Papers)}, pages 4781--4791,
  Online. Association for Computational Linguistics.

\bibitem[{Damonte and Cohen(2018)}]{damonte-cohen-2018-cross}
Marco Damonte and Shay~B. Cohen. 2018.
\newblock \href {https://doi.org/10.18653/v1/N18-1104} {Cross-lingual
  {A}bstract {M}eaning {R}epresentation parsing}.
\newblock In \emph{Proceedings of the 2018 Conference of the North {A}merican
  Chapter of the Association for Computational Linguistics: Human Language
  Technologies, Volume 1 (Long Papers)}, pages 1146--1155, New Orleans,
  Louisiana. Association for Computational Linguistics.

\bibitem[{Deshpande and Narasimhan(2020)}]{deshpande-narasimhan-2020-guiding}
Ameet Deshpande and Karthik Narasimhan. 2020.
\newblock \href {https://doi.org/10.18653/v1/2020.findings-emnlp.419} {Guiding
  attention for self-supervised learning with transformers}.
\newblock In \emph{Findings of the Association for Computational Linguistics:
  EMNLP 2020}, pages 4676--4686, Online. Association for Computational
  Linguistics.

\bibitem[{Dou and Neubig(2021)}]{dou-neubig-2021-word}
Zi-Yi Dou and Graham Neubig. 2021.
\newblock \href {https://doi.org/10.18653/v1/2021.eacl-main.181} {Word
  alignment by fine-tuning embeddings on parallel corpora}.
\newblock In \emph{Proceedings of the 16th Conference of the European Chapter
  of the Association for Computational Linguistics: Main Volume}, pages
  2112--2128, Online. Association for Computational Linguistics.

\bibitem[{Flanigan et~al.(2014)Flanigan, Thomson, Carbonell, Dyer, and
  Smith}]{flanigan-etal-2014-discriminative}
Jeffrey Flanigan, Sam Thomson, Jaime Carbonell, Chris Dyer, and Noah~A. Smith.
  2014.
\newblock \href {https://doi.org/10.3115/v1/P14-1134} {A discriminative
  graph-based parser for the {A}bstract {M}eaning {R}epresentation}.
\newblock In \emph{Proceedings of the 52nd Annual Meeting of the Association
  for Computational Linguistics (Volume 1: Long Papers)}, pages 1426--1436,
  Baltimore, Maryland. Association for Computational Linguistics.

\bibitem[{Hardy and Vlachos(2018)}]{hardy-vlachos-2018-guided}
Hardy Hardy and Andreas Vlachos. 2018.
\newblock \href {https://doi.org/10.18653/v1/D18-1086} {Guided neural language
  generation for abstractive summarization using {A}bstract {M}eaning
  {R}epresentation}.
\newblock In \emph{Proceedings of the 2018 Conference on Empirical Methods in
  Natural Language Processing}, pages 768--773, Brussels, Belgium. Association
  for Computational Linguistics.

\bibitem[{Kapanipathi et~al.(2021)Kapanipathi, Abdelaziz, Ravishankar, Roukos,
  Gray, Fernandez~Astudillo, Chang, Cornelio, Dana, Fokoue, Garg, Gliozzo,
  Gurajada, Karanam, Khan, Khandelwal, Lee, Li, Luus, Makondo,
  Mihindukulasooriya, Naseem, Neelam, Popa, Gangi~Reddy, Riegel, Rossiello,
  Sharma, Bhargav, and Yu}]{kapanipathi-etal-2021-leveraging}
Pavan Kapanipathi, Ibrahim Abdelaziz, Srinivas Ravishankar, Salim Roukos,
  Alexander Gray, Ram{\'o}n Fernandez~Astudillo, Maria Chang, Cristina
  Cornelio, Saswati Dana, Achille Fokoue, Dinesh Garg, Alfio Gliozzo, Sairam
  Gurajada, Hima Karanam, Naweed Khan, Dinesh Khandelwal, Young-Suk Lee, Yunyao
  Li, Francois Luus, Ndivhuwo Makondo, Nandana Mihindukulasooriya, Tahira
  Naseem, Sumit Neelam, Lucian Popa, Revanth Gangi~Reddy, Ryan Riegel, Gaetano
  Rossiello, Udit Sharma, G~P~Shrivatsa Bhargav, and Mo~Yu. 2021.
\newblock \href {https://doi.org/10.18653/v1/2021.findings-acl.339} {Leveraging
  {A}bstract {M}eaning {R}epresentation for knowledge base question answering}.
\newblock In \emph{Findings of the Association for Computational Linguistics:
  ACL-IJCNLP 2021}, pages 3884--3894, Online. Association for Computational
  Linguistics.

\bibitem[{Kokhlikyan et~al.(2020)Kokhlikyan, Miglani, Martin, Wang, Alsallakh,
  Reynolds, Melnikov, Kliushkina, Araya, Yan, and
  Reblitz-Richardson}]{kokhlikyan2020captum}
Narine Kokhlikyan, Vivek Miglani, Miguel Martin, Edward Wang, Bilal Alsallakh,
  Jonathan Reynolds, Alexander Melnikov, Natalia Kliushkina, Carlos Araya, Siqi
  Yan, and Orion Reblitz-Richardson. 2020.
\newblock \href {http://arxiv.org/abs/2009.07896} {Captum: A unified and
  generic model interpretability library for pytorch}.

\bibitem[{Lewis et~al.(2020)Lewis, Liu, Goyal, Ghazvininejad, Mohamed, Levy,
  Stoyanov, and Zettlemoyer}]{lewis-etal-2020-bart}
Mike Lewis, Yinhan Liu, Naman Goyal, Marjan Ghazvininejad, Abdelrahman Mohamed,
  Omer Levy, Veselin Stoyanov, and Luke Zettlemoyer. 2020.
\newblock \href {https://doi.org/10.18653/v1/2020.acl-main.703} {{BART}:
  Denoising sequence-to-sequence pre-training for natural language generation,
  translation, and comprehension}.
\newblock In \emph{Proceedings of the 58th Annual Meeting of the Association
  for Computational Linguistics}, pages 7871--7880, Online. Association for
  Computational Linguistics.

\bibitem[{Liao et~al.(2018)Liao, Lebanoff, and Liu}]{liao-etal-2018-abstract}
Kexin Liao, Logan Lebanoff, and Fei Liu. 2018.
\newblock \href {https://aclanthology.org/C18-1101} {{A}bstract {M}eaning
  {R}epresentation for multi-document summarization}.
\newblock In \emph{Proceedings of the 27th International Conference on
  Computational Linguistics}, pages 1178--1190, Santa Fe, New Mexico, USA.
  Association for Computational Linguistics.

\bibitem[{Lim et~al.(2020)Lim, Oh, Jang, Yang, and Lim}]{lim-etal-2020-know}
Jungwoo Lim, Dongsuk Oh, Yoonna Jang, Kisu Yang, and Heuiseok Lim. 2020.
\newblock \href {https://doi.org/10.18653/v1/2020.coling-main.222} {{I} know
  what you asked: Graph path learning using {AMR} for commonsense reasoning}.
\newblock In \emph{Proceedings of the 28th International Conference on
  Computational Linguistics}, pages 2459--2471, Barcelona, Spain (Online).
  International Committee on Computational Linguistics.

\bibitem[{Liu et~al.(2018)Liu, Che, Zheng, Qin, and Liu}]{liu-etal-2018-amr}
Yijia Liu, Wanxiang Che, Bo~Zheng, Bing Qin, and Ting Liu. 2018.
\newblock \href {https://doi.org/10.18653/v1/D18-1264} {An {AMR} aligner tuned
  by transition-based parser}.
\newblock In \emph{Proceedings of the 2018 Conference on Empirical Methods in
  Natural Language Processing}, pages 2422--2430, Brussels, Belgium.
  Association for Computational Linguistics.

\bibitem[{Liu et~al.(2020)Liu, Gu, Goyal, Li, Edunov, Ghazvininejad, Lewis, and
  Zettlemoyer}]{liu-etal-2020-multilingual-denoising}
Yinhan Liu, Jiatao Gu, Naman Goyal, Xian Li, Sergey Edunov, Marjan
  Ghazvininejad, Mike Lewis, and Luke Zettlemoyer. 2020.
\newblock \href {https://doi.org/10.1162/tacl_a_00343} {Multilingual denoising
  pre-training for neural machine translation}.
\newblock \emph{Transactions of the Association for Computational Linguistics},
  8:726--742.

\bibitem[{Manakul and Gales(2021)}]{manakul-gales-2021-long}
Potsawee Manakul and Mark Gales. 2021.
\newblock \href {https://doi.org/10.18653/v1/2021.acl-long.470} {Long-span
  summarization via local attention and content selection}.
\newblock In \emph{Proceedings of the 59th Annual Meeting of the Association
  for Computational Linguistics and the 11th International Joint Conference on
  Natural Language Processing (Volume 1: Long Papers)}, pages 6026--6041,
  Online. Association for Computational Linguistics.

\bibitem[{Mart{\'\i}nez~Lorenzo et~al.(2022)Mart{\'\i}nez~Lorenzo, Maru, and
  Navigli}]{martinez-lorenzo-etal-2022-fully}
Abelardo~Carlos Mart{\'\i}nez~Lorenzo, Marco Maru, and Roberto Navigli. 2022.
\newblock \href {https://doi.org/10.18653/v1/2022.acl-long.121}
  {{F}ully-{S}emantic {P}arsing and {G}eneration: the {B}abel{N}et {M}eaning
  {R}epresentation}.
\newblock In \emph{Proceedings of the 60th Annual Meeting of the Association
  for Computational Linguistics (Volume 1: Long Papers)}, pages 1727--1741,
  Dublin, Ireland. Association for Computational Linguistics.

\bibitem[{Martins and Astudillo(2016)}]{10.5555/3045390.3045561}
Andr\'{e} F.~T. Martins and Ram\'{o}n~F. Astudillo. 2016.
\newblock \href {http://proceedings.mlr.press/v48/martins16.html} {From softmax
  to sparsemax: A sparse model of attention and multi-label classification}.
\newblock In \emph{Proceedings of the 33rd International Conference on
  International Conference on Machine Learning - Volume 48}, ICML'16, page
  1614–1623. JMLR.org.

\bibitem[{Navigli et~al.(2022)Navigli, Blloshmi, and
  Martinez~Lorenzo}]{bmr-etal-2022-bmr}
Roberto Navigli, Rexhina Blloshmi, and Abelardo~Carlos Martinez~Lorenzo. 2022.
\newblock \href
  {https://www.researchgate.net/publication/358818124_BabelNet_Meaning_Representation_A_Fully_Semantic_Formalism_to_Overcome_Language_Barriers}
  {{BabelNet Meaning Representation: A Fully Semantic Formalism to Overcome
  Language Barriers}}.
\newblock \emph{Proceedings of the AAAI Conference on Artificial Intelligence},
  36.

\bibitem[{Oral and Eryi{\u{g}}it(2022)}]{oral-eryigit-2022-amr}
K.~Elif Oral and G{\"u}l{\c{s}}en Eryi{\u{g}}it. 2022.
\newblock \href {https://doi.org/10.18653/v1/2022.acl-srw.13} {{AMR} alignment
  for morphologically-rich and pro-drop languages}.
\newblock In \emph{Proceedings of the 60th Annual Meeting of the Association
  for Computational Linguistics: Student Research Workshop}, pages 143--152,
  Dublin, Ireland. Association for Computational Linguistics.

\bibitem[{Peters et~al.(2018)Peters, Neumann, Iyyer, Gardner, Clark, Lee, and
  Zettlemoyer}]{peters-etal-2018-deep}
Matthew~E. Peters, Mark Neumann, Mohit Iyyer, Matt Gardner, Christopher Clark,
  Kenton Lee, and Luke Zettlemoyer. 2018.
\newblock \href {https://doi.org/10.18653/v1/N18-1202} {Deep contextualized
  word representations}.
\newblock In \emph{Proceedings of the 2018 Conference of the North {A}merican
  Chapter of the Association for Computational Linguistics: Human Language
  Technologies, Volume 1 (Long Papers)}, pages 2227--2237, New Orleans,
  Louisiana. Association for Computational Linguistics.

\bibitem[{Pourdamghani et~al.(2014)Pourdamghani, Gao, Hermjakob, and
  Knight}]{pourdamghani-etal-2014-aligning}
Nima Pourdamghani, Yang Gao, Ulf Hermjakob, and Kevin Knight. 2014.
\newblock \href {https://doi.org/10.3115/v1/D14-1048} {Aligning {E}nglish
  strings with {A}bstract {M}eaning {R}epresentation graphs}.
\newblock In \emph{Proceedings of the 2014 Conference on Empirical Methods in
  Natural Language Processing ({EMNLP})}, pages 425--429, Doha, Qatar.
  Association for Computational Linguistics.

\bibitem[{Rao et~al.(2017)Rao, Marcu, Knight, and
  Daum{\'e}~III}]{rao-etal-2017-biomedical}
Sudha Rao, Daniel Marcu, Kevin Knight, and Hal Daum{\'e}~III. 2017.
\newblock \href {https://doi.org/10.18653/v1/W17-2315} {Biomedical event
  extraction using {A}bstract {M}eaning {R}epresentation}.
\newblock In \emph{{B}io{NLP} 2017}, pages 126--135, Vancouver, Canada,.
  Association for Computational Linguistics.

\bibitem[{Shrikumar et~al.(2017)Shrikumar, Greenside, and
  Kundaje}]{10.5555/3305890.3306006}
Avanti Shrikumar, Peyton Greenside, and Anshul Kundaje. 2017.
\newblock \href
  {http://proceedings.mlr.press/v70/shrikumar17a/shrikumar17a.pdf} {Learning
  important features through propagating activation differences}.
\newblock In \emph{Proceedings of the 34th International Conference on Machine
  Learning - Volume 70}, ICML'17, page 3145–3153. JMLR.org.

\bibitem[{Simonyan et~al.(2014)Simonyan, Vedaldi, and
  Zisserman}]{Simonyan2014DeepIC}
Karen Simonyan, Andrea Vedaldi, and Andrew Zisserman. 2014.
\newblock \href {https://arxiv.org/pdf/1312.6034.pdf} {Deep inside
  convolutional networks: Visualising image classification models and saliency
  maps}.
\newblock \emph{CoRR}, abs/1312.6034.

\bibitem[{Song et~al.(2019)Song, Gildea, Zhang, Wang, and
  Su}]{song-etal-2019-semantic}
Linfeng Song, Daniel Gildea, Yue Zhang, Zhiguo Wang, and Jinsong Su. 2019.
\newblock \href {https://doi.org/10.1162/tacl_a_00252} {Semantic neural machine
  translation using {AMR}}.
\newblock \emph{Transactions of the Association for Computational Linguistics},
  7:19--31.

\bibitem[{Sood et~al.(2020)Sood, Tannert, Mueller, and
  Bulling}]{NEURIPS2020_460191c7}
Ekta Sood, Simon Tannert, Philipp Mueller, and Andreas Bulling. 2020.
\newblock \href
  {https://proceedings.neurips.cc/paper/2020/file/460191c72f67e90150a093b4585e7eb4-Paper.pdf}
  {Improving natural language processing tasks with human gaze-guided neural
  attention}.
\newblock In \emph{Advances in Neural Information Processing Systems},
  volume~33, pages 6327--6341. Curran Associates, Inc.

\bibitem[{Springenberg et~al.(2015)Springenberg, Dosovitskiy, Brox, and
  Riedmiller}]{Springenberg2015StrivingFS}
Jost~Tobias Springenberg, Alexey Dosovitskiy, Thomas Brox, and Martin~A.
  Riedmiller. 2015.
\newblock \href {http://arxiv.org/abs/1412.6806} {Striving for simplicity: The
  all convolutional net}.
\newblock In \emph{3rd International Conference on Learning Representations,
  {ICLR} 2015, San Diego, CA, USA, May 7-9, 2015, Workshop Track Proceedings}.

\bibitem[{Stacey et~al.(2021)Stacey, Belinkov, and Rei}]{stacey2022aaai}
Joe Stacey, Yonatan Belinkov, and Marek Rei. 2021.
\newblock \href {https://doi.org/10.48550/ARXIV.2104.08142} {Supervising model
  attention with human explanations for robust natural language inference}.

\bibitem[{Tenney et~al.(2019)Tenney, Das, and Pavlick}]{tenney-etal-2019-bert}
Ian Tenney, Dipanjan Das, and Ellie Pavlick. 2019.
\newblock \href {https://doi.org/10.18653/v1/P19-1452} {{BERT} rediscovers the
  classical {NLP} pipeline}.
\newblock In \emph{Proceedings of the 57th Annual Meeting of the Association
  for Computational Linguistics}, pages 4593--4601, Florence, Italy.
  Association for Computational Linguistics.

\bibitem[{Uhrig et~al.(2021)Uhrig, Garcia, Opitz, and
  Frank}]{uhrig-etal-2021-translate}
Sarah Uhrig, Yoalli Garcia, Juri Opitz, and Anette Frank. 2021.
\newblock \href {https://doi.org/10.18653/v1/2021.iwpt-1.6} {Translate, then
  parse! a strong baseline for cross-lingual {AMR} parsing}.
\newblock In \emph{Proceedings of the 17th International Conference on Parsing
  Technologies and the IWPT 2021 Shared Task on Parsing into Enhanced Universal
  Dependencies (IWPT 2021)}, pages 58--64, Online. Association for
  Computational Linguistics.

\bibitem[{Vashishth et~al.(2019)Vashishth, Upadhyay, Tomar, and
  Faruqui}]{Vashishth2019AttentionIA}
Shikhar Vashishth, Shyam Upadhyay, Gaurav~Singh Tomar, and Manaal Faruqui.
  2019.
\newblock \href
  {http://dblp.uni-trier.de/db/journals/corr/corr1909.html#abs-1909-11218}
  {Attention interpretability across nlp tasks.}
\newblock \emph{CoRR}, abs/1909.11218.

\bibitem[{Vaswani et~al.(2017)Vaswani, Shazeer, Parmar, Uszkoreit, Jones,
  Gomez, Kaiser, and Polosukhin}]{NIPS2017_3f5ee243}
Ashish Vaswani, Noam Shazeer, Niki Parmar, Jakob Uszkoreit, Llion Jones,
  Aidan~N Gomez, \L~ukasz Kaiser, and Illia Polosukhin. 2017.
\newblock \href
  {https://proceedings.neurips.cc/paper/2017/file/3f5ee243547dee91fbd053c1c4a845aa-Paper.pdf}
  {Attention is all you need}.
\newblock In \emph{Advances in Neural Information Processing Systems},
  volume~30. Curran Associates, Inc.

\bibitem[{Wein and Schneider(2021)}]{wein-schneider-2021-classifying}
Shira Wein and Nathan Schneider. 2021.
\newblock \href {https://doi.org/10.18653/v1/2021.law-1.6} {Classifying
  divergences in cross-lingual {AMR} pairs}.
\newblock In \emph{Proceedings of The Joint 15th Linguistic Annotation Workshop
  (LAW) and 3rd Designing Meaning Representations (DMR) Workshop}, pages
  56--65, Punta Cana, Dominican Republic. Association for Computational
  Linguistics.

\bibitem[{Wilcoxon(1945)}]{10.2307/3001968}
Frank Wilcoxon. 1945.
\newblock \href {http://www.jstor.org/stable/3001968} {Individual comparisons
  by ranking methods}.
\newblock \emph{Biometrics Bulletin}, 1(6):80--83.

\bibitem[{Wu et~al.(2020)Wu, Nguyen, and Ong}]{wu-etal-2020-structured}
Zhengxuan Wu, Thanh-Son Nguyen, and Desmond Ong. 2020.
\newblock \href {https://doi.org/10.18653/v1/2020.blackboxnlp-1.24} {Structured
  self-{A}ttention{W}eights encode semantics in sentiment analysis}.
\newblock In \emph{Proceedings of the Third BlackboxNLP Workshop on Analyzing
  and Interpreting Neural Networks for NLP}, pages 255--264, Online.
  Association for Computational Linguistics.

\bibitem[{Xu et~al.(2020)Xu, Li, Yuan, Wu, He, and Zhou}]{xu-etal-2020-self}
Song Xu, Haoran Li, Peng Yuan, Youzheng Wu, Xiaodong He, and Bowen Zhou. 2020.
\newblock \href {https://doi.org/10.18653/v1/2020.acl-main.125} {Self-attention
  guided copy mechanism for abstractive summarization}.
\newblock In \emph{Proceedings of the 58th Annual Meeting of the Association
  for Computational Linguistics}, pages 1355--1362, Online. Association for
  Computational Linguistics.

\bibitem[{Yin et~al.(2021)Yin, Fernandes, Pruthi, Chaudhary, Martins, and
  Neubig}]{yin-etal-2021-context}
Kayo Yin, Patrick Fernandes, Danish Pruthi, Aditi Chaudhary, Andr{\'e} F.~T.
  Martins, and Graham Neubig. 2021.
\newblock \href {https://doi.org/10.18653/v1/2021.acl-long.65} {Do
  context-aware translation models pay the right attention?}
\newblock In \emph{Proceedings of the 59th Annual Meeting of the Association
  for Computational Linguistics and the 11th International Joint Conference on
  Natural Language Processing (Volume 1: Long Papers)}, pages 788--801, Online.
  Association for Computational Linguistics.

\bibitem[{Zeiler and Fergus(2014)}]{Zeiler2014VisualizingAU}
Matthew~D. Zeiler and Rob Fergus. 2014.
\newblock \href {https://cs.nyu.edu/~fergus/papers/zeilerECCV2014.pdf}
  {Visualizing and understanding convolutional networks}.
\newblock In \emph{ECCV}.

\bibitem[{Zhang and Feng(2021)}]{zhang-feng-2021-modeling-concentrated}
Shaolei Zhang and Yang Feng. 2021.
\newblock \href {https://doi.org/10.18653/v1/2021.findings-emnlp.121} {Modeling
  concentrated cross-attention for neural machine translation with {G}aussian
  mixture model}.
\newblock In \emph{Findings of the Association for Computational Linguistics:
  EMNLP 2021}, pages 1401--1411, Punta Cana, Dominican Republic. Association
  for Computational Linguistics.

\bibitem[{Zhou et~al.(2021)Zhou, Naseem, Fernandez~Astudillo, Lee, Florian, and
  Roukos}]{zhou-etal-2021-structure}
Jiawei Zhou, Tahira Naseem, Ram{\'o}n Fernandez~Astudillo, Young-Suk Lee, Radu
  Florian, and Salim Roukos. 2021.
\newblock \href {https://doi.org/10.18653/v1/2021.emnlp-main.507}
  {Structure-aware fine-tuning of sequence-to-sequence transformers for
  transition-based {AMR} parsing}.
\newblock In \emph{Proceedings of the 2021 Conference on Empirical Methods in
  Natural Language Processing}, pages 6279--6290, Online and Punta Cana,
  Dominican Republic. Association for Computational Linguistics.

\end{thebibliography}
\bibliographystyle{acl_natbib}

\appendix

\section{LEAMR Alignment Rules}
\label{sec:rules}
The LEAMR standard has some predefined strategies for alignments that were followed during their annotation, as well as fixed in their alignment pipeline along EM. We kept a few of these strategies when extracting the alignment, just those related to the structure of the graph, but not those concerning token matching between the sentence and the graph.
\subsection{Subgraph} 
\begin{itemize}
    \item Nodes \textit{have-org-role-91} and \textit{have-rel-role-91} follow a fixed structure related to a person ie. the sentence word \textit{enemy} is represented as \textit{person $\rightarrow$ have-rel-role-91 $\rightarrow$ enemy}, therefore for such subgraphs we use the alignment from the child node.
    \item Similarly for Named Entities, we align the whole subgraph structure based on its child nodes which indicate its surfaceform. However this leads to some errors as described in Section \ref{sec:errors}. 
    \item We align node \textit{amr-unknown} to the question mark if it appears in the sentence.
\end{itemize}

\subsection{Relations} 
\begin{itemize}
    \item For the relation \textit{:condition} we align it to the word \textit{if} when it appears in the sentence.
    \item \textit{:purpose} is aligned with \textit{to} when in the sentence.
    \item \textit{:ARGX} relations are aligned to the same span as the parent node, while \textit{:ARGX-of} to that of the child, since they share the alignment of the predicate they are connected to.
    \item For \textit{:mod} and \textit{:duration} we use the alignment from the child node.
    \item For \textit{:domain} and \textit{:opX} we use the alignment from the parent node.
\end{itemize}

\section{Extra Results}
\label{sec:appendix_isi}
\subsection{LEAMR Results}
\begin{table*}[h!]
\centering
\begin{tabular}{ll|ccc|ccc|c}
\toprule
                                &         & \multicolumn{3}{c|}{\textbf{Exact Alignment}}                                                        & \multicolumn{3}{c|}{\textbf{Partial Alignment}}                                             & \multicolumn{1}{l}{\textbf{Spans}} \\
                                &         & \textbf{P}                         & \textbf{R}                                  & \textbf{F1}                         & \textbf{P}                         & \textbf{R}                         & \textbf{F1}                         & \textbf{F1}                        \\  \midrule \midrule
\multicolumn{1}{l|}{\textbf{Subgraph}}   & Run 1   & \textbf{94.39}            & \textbf{94.67}                     & \textbf{94.53}             & \textbf{96.62}            & \textbf{96.90}            & \textbf{96.76}             & \textbf{96.40}            \\
\multicolumn{1}{l|}{\textbf{Alignment}}  & Run 2   & 93.79                     & 93.85                              & 93.82                      & 96.22                     & 96.27                     & 96.25                      & 96.05                     \\
\multicolumn{1}{l|}{(1707)}     & Run 3   & 94.26                     & 94.32                              & 94.29                      & 96.60                     & 96.66                     & 96.63                      & 96.34                     \\
\multicolumn{1}{l|}{}           & Run 4   & 94.20                     & 94.26                              & 94.23                      & 96.47                     & 96.53                     & 96.50                      & 96.22                     \\
\multicolumn{1}{l|}{}           & Run 5   & 93.81                     & 94.14                              & 93.98                      & 95.81                     & 96.14                     & 95.97                      & 95.73                     \\ \midrule
\multicolumn{1}{l|}{}           & Average & 94.09                     & 94.25                              & 94.17                      & 96.34                     & 96.50                     & 96.42                      & 96.15                     \\
\multicolumn{1}{l|}{}           & Std     & 0.27                      & 0.30                               & 0.28                       & 0.34                      & 0.30                      & 0.32                       & 0.27                      \\  \midrule \midrule
\multicolumn{1}{l|}{\textbf{Relation}}   & Run 1   & 88.03                     & 88.18                              & 88.11                      & 91.08                     & 91.24                     & 91.16                      & 91.87                     \\
\multicolumn{1}{l|}{\textbf{Alignment}}  & Run 2   & 87.90                     & 88.36                              & 88.13                      & 90.71                     & 91.18                     & 90.95                      & 91.87                     \\
\multicolumn{1}{l|}{(1263)}     & Run 3   & \textbf{88.61}            & \textbf{88.61}                     & \textbf{88.61}             & \textbf{91.44}            & \textbf{91.44}            & \textbf{91.44}             & \textbf{91.95}            \\
\multicolumn{1}{l|}{}           & Run 4   & 88.39                     & \textbf{88.61}                     & 88.50                      & 91.02                     & 91.25                     & 91.14                      & 91.66                     \\
\multicolumn{1}{l|}{}           & Run 5   & 88.59                     & 88.44                              & 88.52                      & 91.24                     & 91.08                     & 91.16                      & 91.86                     \\ \midrule
\multicolumn{1}{l|}{}           & Average & 88.30                     & 88.44                              & 88.37                      & 91.10                     & 91.24                     & 91.17                      & 91.84                     \\
\multicolumn{1}{l|}{}           & Std     & 0.32                      & 0.18                               & 0.28                       & 0.27                      & 0.13                      & 0.17                       & 0.05                      \\ \midrule \midrule
\multicolumn{1}{l|}{\textbf{Reentrancy}} & Run 1   & 56.90                     & 57.09                              & 57.00                      &           ---                &                      ---     &          ---                  &                      ---     \\
\multicolumn{1}{l|}{\textbf{Alignment}}  & Run 2   & 56.23                     & 56.42                              & 56.32                      &               ---            &                      ---     &             ---               &                     ---      \\
\multicolumn{1}{l|}{(293)}      & Run 3   & \textbf{57.24}                     & \textbf{57.43}                              & \textbf{57.34}                      &     ---                      &                ---           &         ---                   &          ---                 \\
\multicolumn{1}{l|}{}           & Run 4   & 55.56                     & 55.74                              & 55.65                      &                ---           &                      ---     &                    ---        &          ---                 \\
\multicolumn{1}{l|}{}           & Run 5   & \multicolumn{1}{l}{55.22} & \multicolumn{1}{l}{55.41}          & \multicolumn{1}{l|}{55.31} & ---   & ---  & ---     & ---\\ \midrule
\multicolumn{1}{l|}{}           & Average & \multicolumn{1}{l}{56.23} & \multicolumn{1}{l}{56.42}          & \multicolumn{1}{l|}{56.32} & ---     & ---     & ---      & ---     \\
\multicolumn{1}{l|}{}           & Std     & 0.86                      & 0.86                               & 0.86                       &                   ---        &                    ---       &           ---                 &                   ---        \\  \midrule \midrule
\multicolumn{1}{l|}{\textbf{Duplicate}}  & Run 1   & 70.00                     & \textbf{82.35}                     & 75.88                      & 72.50                     & 85.29                     & 78.38                      &                 ---          \\
\multicolumn{1}{l|}{\textbf{Subgraph}}   & Run 2   & 65.00                     & 76.47                              & 70.27                      & 67.50                     & 79.41                     & 72.97                      &               ---            \\
\multicolumn{1}{l|}{\textbf{Alignment}}  & Run 3   & 70.00                     & \textbf{82.35}                     & 75.68                      & 70.00                     & 82.35                     & 75.68                      &            ---               \\
\multicolumn{1}{l|}{(17)}       & Run 4   & \textbf{73.68}            & \textbf{82.35}                     & \textbf{77.78}             & \textbf{76.32}            & \textbf{85.29}            & \textbf{80.56}             &                ---           \\
\multicolumn{1}{l|}{}           & Run 5   & \multicolumn{1}{l}{70.00} & \multicolumn{1}{l}{\textbf{82.35}} & \multicolumn{1}{l|}{75.68} & \multicolumn{1}{l}{70.00} & \multicolumn{1}{l}{82.35} & \multicolumn{1}{l|}{75.68} & ---  \\ \midrule
\multicolumn{1}{l|}{}           & Average & \multicolumn{1}{l}{69.74} & \multicolumn{1}{l}{81.17}          & \multicolumn{1}{l|}{75.06} & \multicolumn{1}{l}{71.26} & \multicolumn{1}{l}{82.94} & \multicolumn{1}{l|}{76.65} & ---     \\
\multicolumn{1}{l|}{}           & Std     & 3.09                      & 2.63                               & 2.82                       & 3.33                      & 2.46                      & 2.90                       &      ---                     \\  \bottomrule
\end{tabular}
\caption{Results on the LEAMR alignment for 5 seeds on the guided approach. Column blocks: runs; measures. Row blocks: alignment types; average and standard deviation (std). Bold is best.}

\label{tab:leamr-runs}
\end{table*}

We explore the variance with different seeds when guiding cross-attention. Table \ref{tab:leamr-experiment} reports on a single seed selected at random. Table \ref{tab:leamr-runs} shows the results for five different seeds as well as the average and standard deviation. We observe some variance, especially for those alignment types with fewer elements; however, average performance is always higher than previous approaches.

\section{Alignment Extraction Algorithm} \label{sec:alignment-extraction-algorithm}

Algorithm \ref{alg:pseudo} shows the procedure for extracting the alignment between spans in the sentence and the semantic units
in the graphs, using a matrix that weights Encoder tokens with the Decoder tokens 

\begin{algorithm*}[t!] 
	\begin{algorithmic}[1]
        
	    \Function{ExtractAlignments}{$encoderTokens, DecoderTokens, scoreMatrix$}
        
        \State $alignmentMap \gets dict()$
        \State $spansList \gets \Call{Spans}{encoderTokens}$\Comment{Extract sentence spans as in LEAMR}
        \State $spanPosMap \gets \Call{tok2span}{encoderTokens}$\Comment{Map input tokens to spans}
        \State $graphPosMap \gets \Call{tok2node}{DecoderTokens}$\Comment{Map output tokens to graph unit}

        \State $\Call{CombineSubwordTokens}{scoreMatrix}$
        
      	\For {$DecoderTokenPos, GraphUnit$ in $graphPosMap$}
        \State $encoderTokensScores \gets scoreMatrix[DecoderTokenPos]$
        \State $maxScorePos \gets \Call{argmax}{encoderTokensScores}$
        \State $alignmentMap[GraphUnit] \gets \Call{SelectSpan}{spansList, maxScorePos}$
		\EndFor
	
        \State $fixedMatches \gets \Call{getFixedMatches}{graphPosMap}$ \Comment{Look for rule based matches} 
        \State $alignmentMap \gets \Call{applyFixedMatches}{alignmentMap, fixedMatches}$ 
		
	    \State $alignments \gets \Call{FormatAlignment}{alignmentMap}$

		\State \Return $alignments$
		\EndFunction
        
	\end{algorithmic}
        
	\caption{Procedure for extracting the alignment between spans in the sentence and the semantic units in the graphs, using a matrix that weights Encoder tokens with the Decoder tokens.}
        \label{alg:pseudo}
\end{algorithm*} 

\section{AMR parsing} \label{sec:appendix_parsing}

Since our guided approach was trained with a different loss than the SPRING model, it could influence the performance in the Semantic Parsing task. Therefore, we also tested our model in the AMR parsing task using the test set of AMR 2.0 and AMR 3.0. Table \ref{tab:amr-parsing} shows the result, where we can observe how our model preserves the performance on parsing.

\begin{table}[ht!]
\centering
\begin{tabular}{l|c|c}
\toprule
                      & \multicolumn{1}{l}{AMR 2.0} & \multicolumn{1}{l}{AMR 3.0}  \\ \midrule \midrule
SPRING                & 84.3  & 83.0                       \\
Ours - Guided - ISI   & 84.3   & 83.0                      \\
Ours - Guided - Leamr & 84.3   & 83.0  \\
\bottomrule
\end{tabular}
\caption{AMR parsing Results.}

\label{tab:amr-parsing}
\end{table}

\section{Hardware}
\label{sec:hardware}
Experiments were performed using a single NVIDIA 3090 GPU with 64GB of RAM and Intel\textsuperscript{\textregistered} Core\textsuperscript{\texttrademark} i9-10900KF CPU.

Training the model took 13 hours, 30 min per training epoch while evaluating on the validation set took 20 min at the end of each epoch. We selected the best performing epoch based on the SMATCH metric on the validation set.

\section{Data}
The AMR data used in this paper is licensed under the \textit{LDC User Agreement for Non-Members} for LDC subscribers, which can be found \href{https://catalog.ldc.upenn.edu/LDC2020T02}{here}. The \textit{The Little Prince} Corpus can be found \href{https://amr.isi.edu/download.html}{here} from the Information Science Institute of the University of Southern California.

\section{Limitations}
Even though our method is an excellent alternative to the current AMR aligner system, which is standard and task-agnostic, we notice some drawbacks when moving to other autoregressive models or languages:

\textbf{Model} $\quad$ In this work, we studied how Cross Attention layers retain alignment information between input and output tokens in auto-regressive models. In Section \ref{sec:experiments}, we examined which layers in state-of-the-art AMR parser models based on BART-large best preserve this information.  Unfortunately, we cannot guarantee that these layers are optimal for other auto-regressive models, and so on. As a result, an examination of cross-attention across multiple models should be done before developing the cross-lingual application of this approach.

\textbf{Sentence Segmentation} $\quad$ It is necessary to apply LEAMR's Spam Segmentation technique to produce the alignment in LEAMR format (Section \ref{sec:align_ext}). However, this segmentation method has several flaws: i) As stated in Section \ref{sec:errors}, this approach does not deal appropriately with phrasal verbs and consecutive segments; ii) the algorithm is English-specific; it is dependent on English grammar rules that we are unable to project to other languages. Therefore we cannot extract the LEAMR alignments in a cross-lingual AMR parsing because we lack a segmentation procedure. However, although LEAMR alignment has this constraint, ISI alignment does not require any initial sentence segmentation and may thus be utilized cross-lingually.

\end{document}